\documentclass[10pt,journal,compsoc]{IEEEtran}

\usepackage{cite}
\usepackage{amsmath,amssymb,amsfonts}
\usepackage{algorithmic}
\usepackage{graphicx}
\usepackage{textcomp}
\usepackage{xcolor}
\usepackage{enumitem}
\usepackage{booktabs}
\usepackage{makecell}
\usepackage{xspace}
\usepackage{multirow}
\usepackage{subfig}
\usepackage{url}

\graphicspath{{Images/}}

\newcommand{\sysname}{{STS}\xspace}

\newcommand{\funcone}{F_{MSA}\xspace}
\newcommand{\functwo}{F_{TRR}\xspace}
\newcommand{\functhree}{F_{DSA}\xspace}

\begin{document}

\title{Spatiotemporal and Semantic Zero-inflated Urban Anomaly Prediction}
%
%

\author{Yao~Lu,~
        Pengyuan~Zhou*,~\IEEEmembership{Member,~IEEE,}
        Yong~Liao,
        and~Haiyong~Xie,~\IEEEmembership{Senior~Member,~IEEE}
\thanks{ Yao Lu, Pengyuan Zhou and Yong Liao are with the School of Cyber Science and Technology, University of Science and Technology of China, Hefei, Anhui 230026, China. 
E-mail: luyao1225@mail.ustc.edu.cn, {	pyzhou, yliao}@ustc.edu.cn. Haiyong Xie is with Adv. Innovation Center for Human Brain Protection, Capital Medical University, Fengtai, BeiJing 100071, China.
E-mail: haiyong.xie@ieee.org. \\ Corresponding author: Pengyuan Zhou}
}


\IEEEtitleabstractindextext{%
\begin{abstract}
Urban anomaly predictions, such as traffic accident prediction and crime prediction, are of vital importance to smart city security and maintenance. 
Existing methods typically use deep learning to capture the intra-dependencies in spatial and temporal dimensions. However, numerous key challenges remain unsolved, for instance, sparse zero-inflated data due to urban anomalies occurring with low frequency (which can lead to poor performance on real-world datasets), and both intra- and inter-dependencies of abnormal 
patterns across spatial, temporal, and semantic dimensions. Moreover, a unified approach to predict multiple kinds of anomaly is left to explore.
In this paper, we propose \sysname to jointly capture the intra- and inter-dependencies between the patterns and the influential factors in three dimensions. 
Further, we use a multi-task prediction module with a customized loss function to solve the zero-inflated issue. 
To verify the effectiveness of the model, we apply it to two urban anomaly prediction tasks, crime prediction and traffic accident risk prediction, respectively.
Experiments on two application scenarios with four real-world datasets demonstrate the superiority of \sysname, which outperforms state-of-the-art methods in the mean absolute error and the root mean square error by 37.88\% and 18.10\% on zero-inflated datasets, and, 60.32\% and 37.28\% on non-zero datasets, respectively.
\end{abstract}

}

\maketitle

\IEEEdisplaynontitleabstractindextext
\IEEEpeerreviewmaketitle

\IEEEraisesectionheading{\section{Introduction}\label{sec:introduction}}
\IEEEPARstart{A}{s} the urban population and the size of cities continue to increase, urban anomalies (e.g., traffic accidents, crimes) are on the rise. A shooting spree at Robb Elementary School in Uvalde, Texas, on May 24, 2022, left 21 people dead, including 18 children and 3 adults. According to the "Gun Violence Archive" website\footnote{\url{https://www.gunviolencearchive.org}}, as of June 15, 19,844 people had been killed and 16,800 had been injured in the United States since 2022 as a result of gun violence. If not handled correctly and promptly, urban abnormalities may also have more severe repercussions, according to~\cite{2018Detecting}. On January 26, 2017, a traffic accident in Harbin led to a chain of rear-end collisions that left 32 people injured and 8 dead. 

If urban anomalies in a city can be quickly detected and even accurately predicted, it can greatly aid the government and related institutions in making integrated planning in advance, allocating resources (such as police resources) fairly, preventing the occurrence of abnormal events with harmful effects, and responding quickly and effectively to reduce the subsequent harm.
For example, crime prediction can help avoid the severe impacts of crimes on society and is thus critical for social well-being. It plays a key role in criminology as an essential prerequisite for crime prevention. 
In particular, some countries have established specialized crime forecasting agencies and law enforcement departments that can use crime forecasting to develop appropriate countermeasures. For instance, Chen et al. designed a routing~\cite{2017Developing} and a districting~\cite{2019Designing} strategy for the police patrol tasks based on crime prediction results. 
Despite the effort, the yearly number of urban anomalies continues to increase, indicating the demand for better prediction methodologies to help prevent urban anomalies in practice. 


Urban anomaly prediction judges the status and development of potential urban anomalies in the future via the analysis of various relevant factors using sensory data, thus promoting the construction of smart cities. 
In order to facilitate urban anomaly prediction, traditional methods typically employ simple mathematical procedures and require additional regional information such as population data (e.g., demographics, poverty ratio, and education level), geographic data, weather data, and anomalies frequency.
However, such methods lack the capability to capture the inherent but intricate temporal and spatial features of criminal events; hence, they are not likely to produce sufficiently accurate predictions.
In recent years, urban anomaly prediction has advanced significantly, thanks to rapidly evolving deep learning methods, particularly in spatiotemporal sequence prediction.
For example, many deep learning-based methods use graph neural networks (GNN) and recurrent neural networks (RNN) to capture spatial and temporal correlation, respectively.

Numerous key challenges remain in the deep learning-based approaches. 
Firstly, the occurrence of anomalies 
is a rare case compared to other time-series events, thus resulting in the so-called zero-inflated issue, namely, the data is extremely sparse and there are many zeros in the collected data. Models trained with zero-inflated data without proper processes tend to generate 0 as the prediction results. This issue severely impacts effectiveness; however, it has been largely overlooked in recent works. 
Secondly, urban anomaly prediction is jointly impacted by the factors from aforementioned dimensions. Most current methods formulate the influence by simply combining the intra-dimension correlations from different dimensions without considering the joint inter-dimension correlations. 
%

To tackle these challenges, we propose a novel framework, \sysname, which jointly models the spatiotemporal and semantic intra- and inter-dependencies, in order to guarantee superior urban anomaly prediction performance. \sysname consists of three key designs: first, \sysname iterates a spatiotemporal and semantic dependency layer, namely STC layer, to jointly capture the dependencies of different categories 
and temporal relationships from nearby regions; second, the STC layer employs a dynamic spatiotemporal attention module to capture cross-spatial and long-term temporal correlations; 
finally, a multi-task prediction module with a customized loss function is proposed to alleviate the zero-inflated issue. Our contributions can be summarized as follows:
\begin{itemize}
\item We propose \sysname to forecast urban anomalies. \sysname applies a three-level gated recursive unit (GRU) architecture and the attention mechanism to extract intra-correlations of individual factors affecting the occurrence of urban anomalies. Subsequently, it captures the spatiotemporal correlations between the target region sequences and their neighbor sequences through a designed attention mechanism; and then captures the effects of multi-hop neighbors through neighborhood aggregation of the region graph. As a result, \sysname can jointly model the intra- and inter- correlations of geospatial factors, long-term and short-term temporal factors, and 
semantic factors.  
\item \sysname employs a multi-task prediction module and a customized loss function to solve the pervasive zero-inflated issue in urban anomaly data. The multi-task prediction module sets the main and auxiliary tasks to simulate the counting process and exposure process in urban anomaly prediction, respectively. Then it reduces the weight of zero samples in the loss function to further solve the zero-inflated issue. 
To the best of our knowledge, \sysname is the first deep-learning urban anomaly prediction method that incorporates a dedicated module to solve the zero-inflated issue. 
%
\item We have conducted extensive experiments in two representative urban anomaly cases using real datasets of New York City and Chicago. Compared with the state-of-the-art~(SOTA) methods, \sysname improves the mean absolute error (MAE) and the root mean squared error (RMSE) by 37.88\% and 18.10\% on zero-inflated datasets, and, 60.32\% and 37.28\% on non-zero datasets, respectively. Experiments on datasets with different sparsity rates verify that \sysname generally performs better on sparser data than SOTA methods.
\end{itemize}

The remainder of the paper is organized as follows. Section~\ref{sec:related} discusses the related works. Section~\ref{sec:method} describes the proposed STS model based on the data preprocessing module, the STC module, and the multi-task prediction module. The practical experiments and relevant analyses are presented in Section~\ref{sec:crime} and Section~\ref{sec:traffic}. We compare and contrast two cases of urban anomaly prediction in great detail in Section~\ref{sec:takeaway}. Section~\ref{sec:conclusion} concludes the paper. 

\section{Related Work}
\label{sec:related}

In this section, we discuss existing studies for urban anomaly prediction. The literature can be largely divided into three categories: traditional statistical methods, traditional machine learning methods, and deep learning based methods. The following three subsections summarize the works associated with these three types of methods.

\subsection{Traditional Statistical Methods}
%
Most of such methods employ simple time-series models, such as autoregressive integrated moving average (ARIMA)~\cite{pan2012utilizing} and its variants~\cite{yadav2018crime}, negative binomial regression~\cite{rumi2019crime}, and geographically weighted negative binomial regression (GWNBR)~\cite{wang2017non}, etc. However, these methods are limited to stable time-series data or otherwise cannot capture the regular patterns. 

\subsection{Traditional Machine Learning Methods}
%
More recently, researchers proposed many methods using traditional machine learning techniques, including support vector regression (SVR)~\cite{wang2010predicting}, support vector machine (SVM)~\cite{kianmehr2008effectiveness}, random forest~\cite{2014Once} and so on. These methods usually only capture either spatial or temporal correlation while ignoring the complex spatiotemporal correlation. Therefore, such methods require more effort on detailed feature engineering in order to fit the scenario for urban anomaly prediction.

\subsection{Deep Learning Based Methods}
%
Deep learning has emerged in recent years as a popular direction in urban anomaly prediction thanks to its capability of capturing the complex nonlinear correlations. Deep learning based methods can be further divided into four categories according to the types of the nonlinear correlations they are designed to capture.

The first category is the general deep learning methods which are directly applied to model spatiotemporal patterns. Mary et al.~\cite{shermila2018crime} adopted the artificial neural network (ANN) to detect criminal patterns. Chun et al.~\cite{chun2019crime} applied the deep neural network (DNN) to individual criminal charges. 
However, these methods do not specifically consider the factors that influence urban anomalies occurrences, thus are not likely to maximize the advantages of deep learning.

The second category is the deep learning methods designed to capture the spatial correlations.
Traditional convolution technologies can only deal with Euclidean structure and therefore are not applicable when the regional network has an irregular topology.
However, the emerging GNN can deal with unstructured data, which greatly promotes the model's capability of capturing spatial correlations. 
For instance, ~\cite{wu2019graph,zheng2020gman,yu2017spatio} all use graph neural networks to capture spatial dependencies. However, the effectiveness of GNN-based techniques for time series prediction is unclear due to the fact that time series data typically lacks graphical structures. 

The third category is the deep learning methods designed to capture temporal correlations.
Such methods usually leverage either the RNN (and its variants) or the attention mechanism.
On the one hand, RNN has been adopted by a large body of literature. For example, Wawrzyniak et al.~\cite{wawrzyniak2018data} used the long short-term memory (LSTM) 
for short-term forecasting. Huang et al. proposed DeepCrime~\cite{huang2018deepcrime} and MiST~\cite{huang2019mist} based on GRU and LSTM, respectively. 
A major drawback of such methods is the long running time due to the recursive structure of RNN. A variant of CNN, namely the temporal convolutional network (TCN), achieves better performance within shorter running time. 
Due to its superiority in sequence modeling, both Wang et al.~\cite{wang2020deep} and Wu et al.~\cite{wu2019graph} use TCN to model temporal features in spatiotemporal prediction problems. 
On the other hand, more recent works (e.g., Wang et al.~\cite{wang2021gsnet}) adopt the attention-based network structure 
, which is more efficient and has significant advantages in long sequence prediction. 
Overall, methods that extract only spatial or only temporal features are not likely to perform well on spatiotemporal data prediction.

The fourth category is the methods designed to simultaneously capture the spatial and temporal correlations for integrated spatiotemporal modeling.
For example, Yu et al.~\cite{yu2017spatio} combined graph convolutional layers and convolutional sequence learning layers to model the spatial dependency on the generated geographical map.
CrimeForecaster~\cite{sun2020crimeforecaster} uses GRU to model temporal dependencies and diffusion convolution modules to capture the cross-region dependencies. 
Unfortunately, the common zero-inflated issue in urban anomaly prediction has not been well addressed. 
Sui et al.~\cite{DBLP:journals/corr/abs-2110-01794} proposed a deep learning method to consider data sparsity in urban anomaly prediction, but its effectiveness questionable as it simply uses deep learning methods that are experimentally more suitable for zero-inflated data without a dedicated module for solving this issue. Consequently, its performance on the non-zero dataset is impacted with poor interpretability of zero-inflated urban anomaly prediction. 

\section{Methodology}
\label{sec:method}
In this section, we first formulate the problem and then present \sysname with detailed descriptions.

\subsection{Formulation}

We consider $C$ types of urban anomalies within $N$ urban regions and $T$ time slots.
The urban regions can be treated as a network, which can further be represented by an undirected graph $\mathcal{G}=(\mathcal{V},\mathcal{E},\mathcal{A})$. Note that $\mathcal{V}$ denotes the set of vertices that represents regions $\mathcal{R}=\{r_1,r_2, \cdots,r_N\}$, $\mathcal{E}$ denotes the set of edges representing the spatial adjacency of the vertices, and $\mathcal{A}\in\mathbb{R}^{N\times N}$ denotes the adjacency matrix of $\mathcal{G}$. Note also that $a_{ij}$ in $\mathcal{A}$ is set to 1 if region $r_i$ and region $r_j$ are geographically neighbors or otherwise 0. 
%

We define a metric, urban anomaly index, to assess the severity of urban anomaly in a region at a certain time. For example, in crime prediction, the urban anomaly index is the number of crime occurrences; in traffic accident risk prediction, the urban anomaly index is the risk of traffic accidents. Higher values represent more anomalous events or more serious situations. We denote the urban anomaly index and the predicted urban anomaly index for type $c$ in region $r$ at $t$-th time slot by $X^t_{r,c}$ and $\hat{X}^t_{r,c}$, respectively. The urban anomaly indices of the entire urban area can be denoted by $X\in \mathbb{R}^{N\times T\times C}$.  

\textbf{Problem Statement.} Given the urban anomaly indices \textit{$X=(X^1, \cdots,X^T)$} in historical $T$ time slots as the input, our goal is to learn a function $F_{\sysname}$ to predict the urban anomaly indices for all urban regions and categories at the next time slot $T+1$: \begin{equation}
    \hat{X}^{T+1} = F_{\sysname}(X^1, \cdots, X^T),
    \nonumber
\end{equation}
where $X^1, \cdots, X^T, \hat{X}^{T+1} \in \mathbb{R}^{N\times C}$.

\begin{figure*}[t!] 
    \centering
    \includegraphics[width=\linewidth]{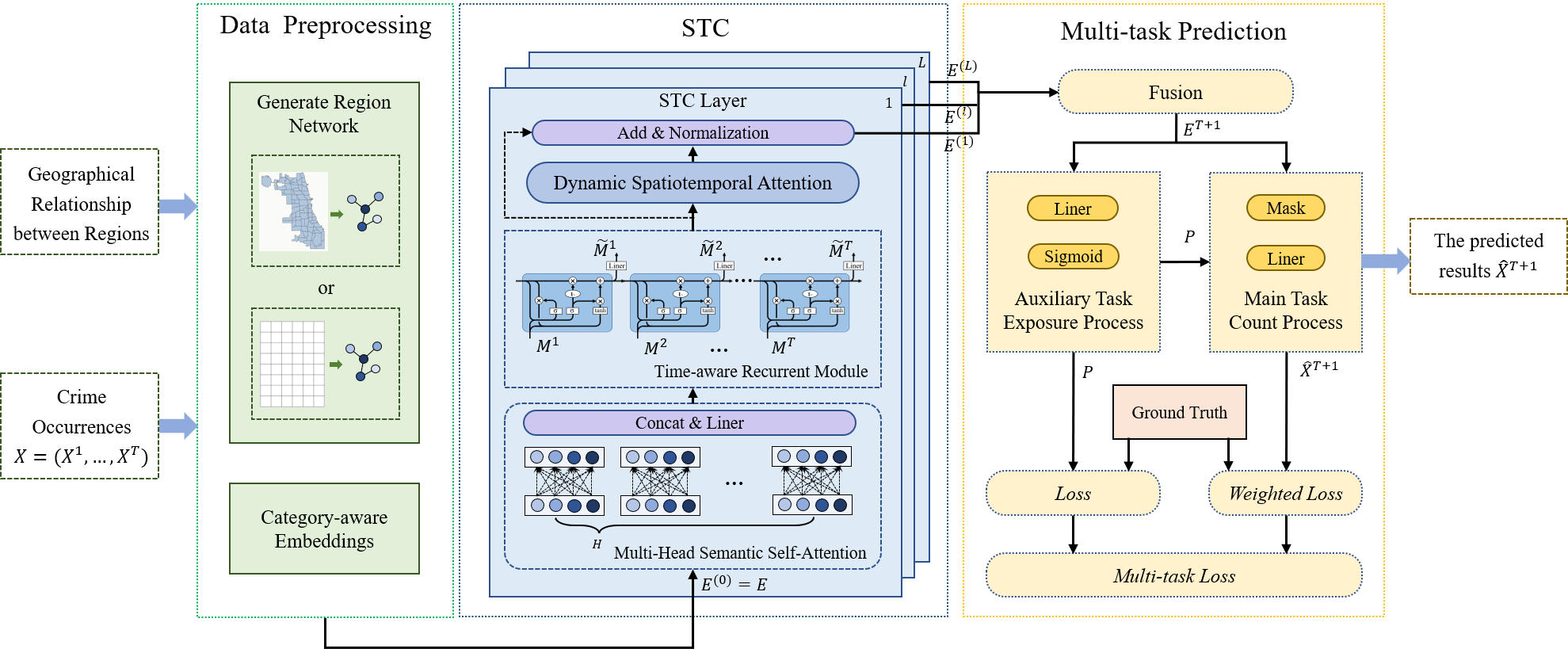}
    \vspace{-0.1in}
    \caption{\sysname\ Framework.$E^{(l)}$ denotes the output of the $l$-th STC layer. $M^t$ and $\widetilde{M}^t$ represent the outputs of the multi-head semantic self-attention and the time-aware recurrent modules in the current STC layer, respectively. $E^{T+1}$ denotes the embeddings at the next time slot $T+1$. $P$ represents the predicted result of the auxiliary task.} 
    \label{fig:model}
    \vspace{-0.1in}
\end{figure*}

\subsection{\sysname Architecture}

\sysname is comprised of three key modules: data preprocessing, STC, and multi-task prediction, as shown in Figure~\ref{fig:model}. 

The data preprocessing module is responsible for partitioning the dataset, generating the region network, and embedding urban anomaly indices with category awareness. 

The STC Module takes as input the embedded urban anomaly indices from the prior $T$ time slots and extracts both spatiotemporal and semantic dependencies simultaneously. 
More specifically, STC stacks multiple STC layers in order to jointly capture the complex correlations between different categories of a urban anomaly (see, e.g.,~\cite{huang2018deepcrime}), among adjacent periods for both long-term temporal dependencies (see, e.g.,~\cite{2018An,DBLP:conf/cikm/ZhaoT17}) and short-term. 

The multi-task prediction module, which is inspired by zero-inflated Poisson regression~\cite{lichman2018prediction} and incorporates a weighted regression loss function, is employed to reduce the negative impacts of zero-inflated data on deep model learning.

\begin{figure}[!t]
\centering
\subfloat[Administrative-region-based]{\includegraphics[width=0.45\linewidth]{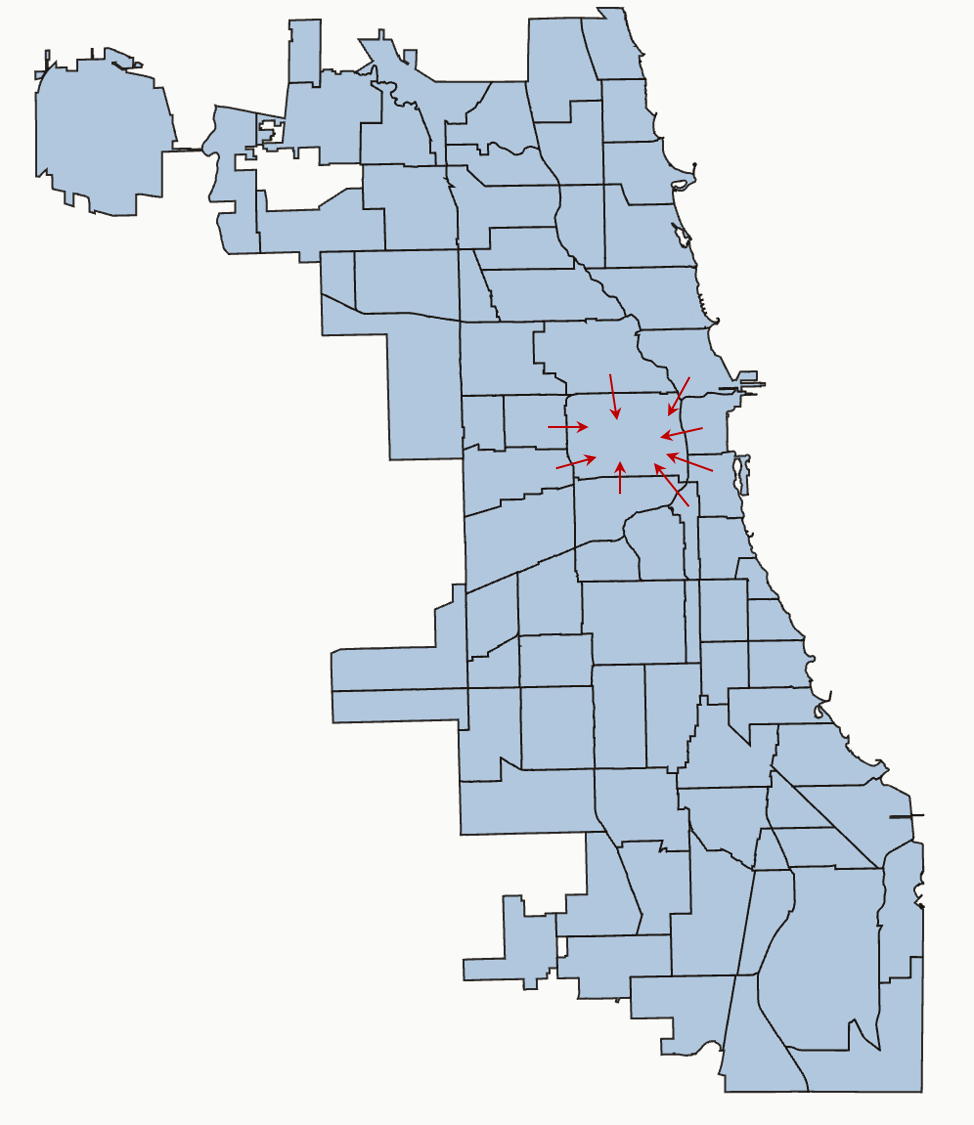}%
\label{fig:dicidation-a}}
\hfil
\subfloat[Administrative-grid-based]{\includegraphics[width=0.45\linewidth]{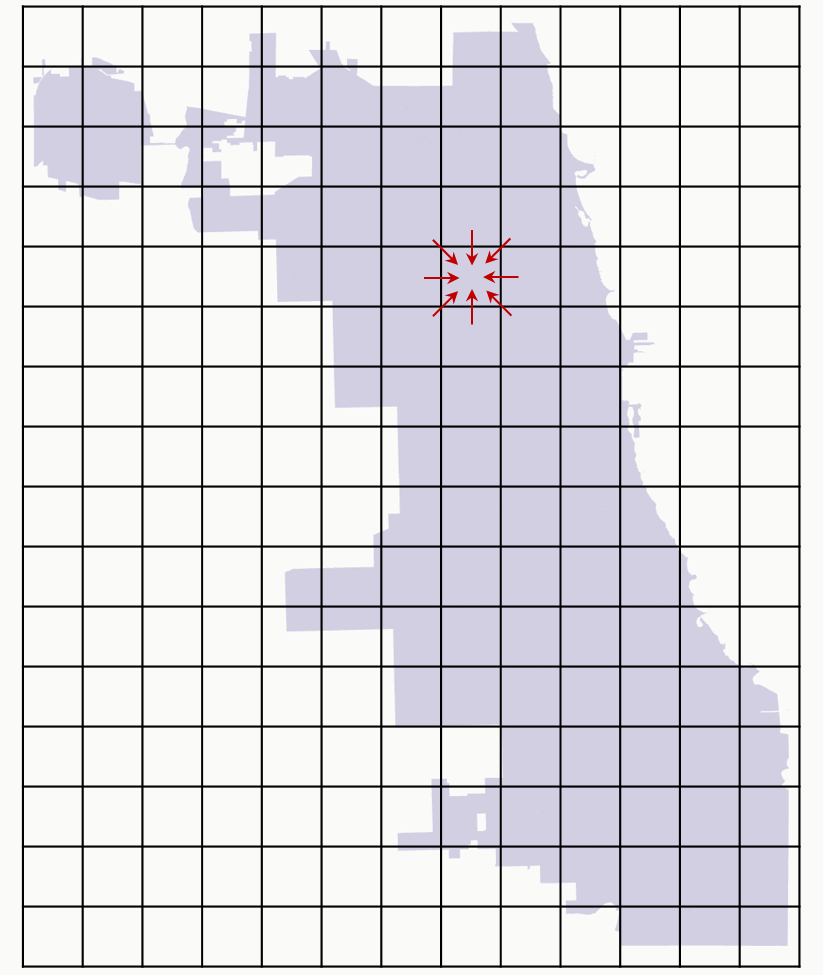}%
\label{fig:dicidation-b}}
\caption{Two Methods of Zoning.}
\label{fig:dicidation}
\end{figure}

\subsection{Data Preprocessing}\label{sec:DP}

Urban anomalies datasets usually partition geographical regions according to either administrative regions or grids.
In order to make \sysname perform well on both partitioning approaches, we transform urban anomaly indices into category-aware embeddings via two strategies for generating the region network $\mathcal{G}$, namely, the administrative-region-based and the grid-based mapping strategies, as shown in Figure~\ref{fig:dicidation}.

More specifically, we denote by $\mathcal{N}_{r_i}$ the adjacent regions (neighbors) of $r_i$, and $s_{i,j}$ the minimum distance between regions $r_i$ and $r_j$, respectively. The adjacency matrix $a_{ij}$ of $r_i$ can be formulated: 
\begin{equation}
  a_{ij} = \begin{cases}
    1 ,s_{i,j}=0 ; \\
    0 ,otherwise.
  \end{cases}
  \nonumber
\end{equation}


We transform urban anomaly indices $X$ into category-aware embeddings $E$~\cite{xia21spatial}, to facilitate extraction of the inter-correlations of cross-category anomalies in dynamic environments. 
In particular, we first normalize the urban anomaly indices $X^t_r$ as follows: \begin{equation}
    \bar{X}^t_r = (X^t_r-\mu)/\sigma,
    \nonumber
\end{equation}
where $\mu$ and $\sigma\in\mathbb{R}^C$  indicate the mean and standard deviation value across $N$ regions and $T$ time slots. 
Then, the category-aware embeddings are generated: 
\begin{equation}
    E^t_{r,c}=\bar{X}^t_{r,c}\cdot e_c,
    \nonumber
\end{equation}
where $e_c\in\mathbb{R}^d$ represents the global embedding vector under the $c$-th category over all time periods, and $d$ is the hidden state size. 
In this way, we obtain an embedding that perceives the category, and the embeddings for each time slot in each region are associated with $C$ types.

\subsection{The STC Module}

The STC module is the main module for capturing urban anomalies data patterns. It takes the output of the data preprocessing module, i.e., $E$, as its input.

The STC module is comprised of multiple stacked STC layers to capture the complex correlations. Each STC layer consists of three components: a multi-head semantic self-attention (MSA), a time-aware recurrent module (TRR), and a dynamic spatiotemporal attention (DSA) component. These components capture the dependencies from one-hop neighbors on semantic, short-term and long-term temporal dimensions, respectively. By stacking multiple STC layers, the impacts of multiple hops away neighbors can be captured. 


%
The $l$-th STC layer computes
\begin{equation}\label{eq:layerC}
    E^{(l)} = \functhree(\functwo(\funcone(E^{(l-1)}))), 
\end{equation}
where $E^{(0)}= E \in\mathbb{R}^{N\times T\times C \times d}$ is the category-aware embeddings generated in data preprocession stage, and $\funcone$, $\functwo$ and $\functhree$ are the functional representations for the three components MSA, TRR and DSA, respectively. 
By composing these functions in the STC layers, we can obtain an embedding that extracts the joint spatiotemporal semantic features. Next we describe the details of the three components of the $l$-th STC layer.

\vskip 0.1in \noindent\textbf{Multi-Head Semantic Self-Attention (MSA).} 
The correlations between different categories of a certain urban anomaly can be arbitrary. Take crime prediction as an example, a robbery that occurred yesterday may reduce the probability of all types of future crime occurring in the region due to increased patrol in the region after the initial robbery~\cite{huang2018deepcrime}. Therefore, a method is needed to model the complex dependencies on the semantic dimension. Considering Transformer's excellent performance in the prediction domain, we employ its core technique, multi-head self-attention, to capture semantic correlations. 
The attention mechanism of each head can capture the representation within a subspace, thus multi-head attention allows the model to focus on information from representation subspaces at different positions~\cite{vaswani2017attention}.

An attention function can be described as mapping a query and a set of key-value pairs to an output. Based on the ($l$-1)-th layer's output embedding $E^{(l-1)}$, we express the query, keys and values in the $h$-th attention head as follows:
\begin{equation}\label{eq:MTC-1}
    \begin{gathered}
        Q^{(l),h}=W^{(l),Q}_h E^{(l-1)}_{c_i},\\
        K^{(l),h}=W^{(l),K}_h E^{(l-1)}_{c_j},\\
        V^{(l),h}=W^{(l),V}_h E^{(l-1)}_{c_j},\\
    \end{gathered}
\end{equation}
where $W^{(l),*}_h\in\mathbb{R}^{d/H\times d}$ indicates the parameters in the $l$-th layer, $H$ is the total number of heads, $E^{(l-1)}_{c_i}$ and $E^{(l-1)}_{c_j}$ denote the embedding of the $c_i$-th and $c_j$-th category after ($l$-1) iterations, respectively.

\par The attention score is calculated by scaled dot-product, which uses highly optimized matrix multiplication to achieve spatial and temporal efficiency. The learnable attention weight $\alpha^{(l),h}_{c_i,c_j}$ and the output $m^{(l),h}_{c_i}$ are generated as follows:
\begin{equation}\label{eq:MTC-2}
\begin{gathered}
        score^{(l),h}_{c_i,c_j}=\frac{{Q^{(l),h}}^\top K^{(l),h}}{\sqrt{d/H}},\\
        \alpha^{(l),h}_{c_i,c_j}=softmax(score^{(l),h}_{c_i,c_j})=\frac{exp(score^{(l),h}_{c_i,c_j})}{\sum_{c_j}exp(score^{(l),h}_{c_i,c_j})},\\
        m^{(l),h}_{c_i}=\sum_{c_j=1}^C\alpha^{(l),h}_{c_i,c_j}\cdot V^{(l),h},
\end{gathered}
\end{equation}
where $d$ is the hidden state size, $H$ is total number of heads, $Q^{(l),h},K^{(l),h},V^{(l),h}$ are query, keys and values calculated by Eq.~\eqref{eq:MTC-1}. 
Let $M^{(l)}_{c_i}$ denote the $c_i$-th type of crime's information gathered from other crime types via the multi-head attention mechanism, which is formally formulated as follows:
\begin{equation}\label{eq:MTC-3}
    M^{(l)}_{c_i}={\big\Arrowvert}^H_{h=1}m^{(l),h}_{c_i}\cdot W^{(l),O},
\end{equation}
where $\big\Arrowvert$ represents the concatenation operation, $W^{(l),O}\in\mathbb{R}^{d\times d}$ is the parameter matrices. We denote $M^{(l)}=\funcone(E^{(l-1)})$, i.e., 
Eq.~\ref{eq:MTC-1}-\ref{eq:MTC-3}, as the function $\funcone$ mentioned in Eq.~\ref{eq:layerC}.


\vskip 0.1in \noindent \textbf{Time-aware Recurrent Module (TRR).} 
There are multiple alternative models that are designed to extract the inter-dependencies in sequences.
For instance, GRU can alleviate the long-term dependency problem of traditional RNN networks. In addition, GRU has similar performance with LSTM but is less computationally expensive.  
Hence, we employ GRU to capture short-term temporal dependencies. Note that GRU cannot focus on too long-term inter-sequence dependencies~\cite{wang2020traffic} and we address the long-term dependencies in the DSA component.

GRU contains a reset gate $r^{(l),t}_{r,c}$ which determines how to integrate the input information with the previous memory, and, an update gate $z^{(l),t}_{r,c}$ which defines the amount of memory preserved to the current time slot. 
Formally, in the $t$-th time slot, the calculation method for hidden states $h^{(l),t}_{r,c}$ and output corresponding to a certain region and a certain category $\widetilde {m}^{(l),t}_{r,c}$ is presented as:
\begin{equation}\label{eq:GRU}
\begin{gathered}
    r^{(l),t}_{r,c}=\sigma(W^{(l)}_r\cdot[h^{(l),t-1}_{r,c},m^{(l),t}_{r,c}]),  \\
    z^{(l),t}_{r,c}=\sigma(W^{(l)}_z\cdot[h^{(l),t-1}_{r,c},m^{(l),t}_{r,c}]),  \\
    \widetilde{h}^{(l),t}_{r,c}=tanh(W^{(l)}_{\widetilde {h}}\cdot[r^{(l),t}_{r,c}*h^{(l),t-1}_{r,c},m^{(l),t}_{r,c}]), \\
    h^{(l),t}_{r,c}=(1-z^{(l),t}_{r,c})* h^{(l),t-1}_{r,c}+z^{(l),t}_{r,c}*\widetilde{h}^{(l),t}_{r,c}, \\
    \widetilde {m}^{(l),t}_{r,c}=\sigma(W^{(l)}_o\cdot h^{(l),t}_{r,c}),
\end{gathered}
\end{equation}
where $W^{(l)}_*\in\mathbb{R}^{d\times d}$ represents the transformation matrix, $\sigma $ represents the sigmoid function, operator * denotes the element-wise product. We denote $\widetilde {M}^{(l)}=\functwo(M^{(l)})$, i.e., Eq.~\ref{eq:GRU} as the function $\functwo$ mentioned in Eq.~\ref{eq:layerC}.


\vskip 0.1in \noindent\textbf{Dynamic Spatiotemporal Attention (DSA).} 
The DSA component targets to extract dependencies between different regions and long-term time frames (note that we capture inter-dependencies across multiple categories and short-term time-series data using the MSA and TRR components, respectively). 

\begin{figure*}
    \centering
    \includegraphics[width=0.8\linewidth]{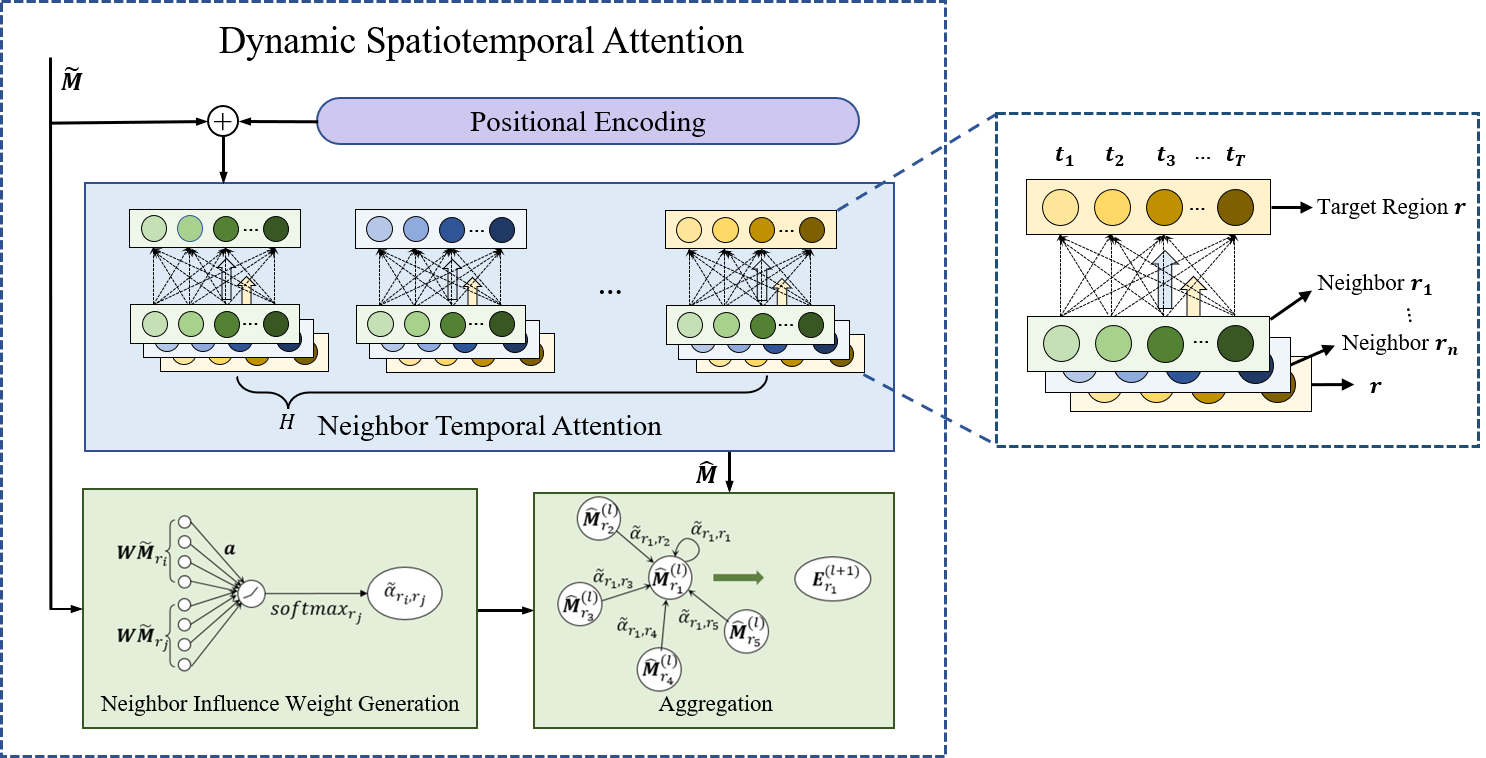}
    \vspace{-0.1in}
    \caption{Dynamic Spatiotemporal Attention}
    \label{fig:DSAttention}
    \vspace{-0.2in}
\end{figure*}

As illustrated in Figure~\ref{fig:DSAttention}, the dynamic spatiotemporal attention firstly uses the output of the time-aware recurrent module $\widetilde {M}^{(l)}$ as the node features, then calculates the influence of neighbors on the target node according to the Graph Attention Networks method, that is, the attention weight $\widetilde\alpha$, which is derived as:
\begin{equation}\label{eq:GAT-1}
    \begin{gathered}
        \widetilde e_{r_i,r_j}=LeakyReLU(a^\top[W^{(l)}\widetilde {M}^{(l)}_{r_i}\big\Arrowvert W^{(l)}\widetilde {M}^{(l)}_{r_j}]), \\
        \widetilde \alpha_{r_i,r_j}=softmax_{r_j}(\widetilde {e}_{r_i,r_j})=\frac{exp(\widetilde e_{r_i,r_j})}{\sum_{r_k\in \mathcal{N}_{r_i}}exp(\widetilde e_{r_i,r_k})}, \\
    \end{gathered}
\end{equation}
where $e_{r_i,r_j}$ denotes the importance of $r_j$ to $r_i$. $W^{(l)}\in\mathbb{R}^{d\times d}$ is the weight matrix parameter that needs to be learned and $a\in\mathbb{R}^{2d}$ is the weight vector of the single-layer feedforward neural network. $\mathcal{N}_{r_i}$ is the neighbors set of region $r_i$. $\widetilde M^{(l)}$ is the output of the time-aware recurrent module. The calculated neighbor influence weight $\widetilde\alpha$ will be used in the final aggregation.


Meanwhile, we design another multi-head self-attention mechanism to achieve the core goal of the DSA module. The input is a time series, thus its order information is a necessity. 
Therefore, prior to the neighbor temporal attention (NTA) module, the positional encoding of the input $\widetilde M^{(l)}$ is required to preserve the sequential information ignored by the attention mechanism. We use the same positional encoding approach as in Transformer. Through the well-designed query, keys and values, we jointly extract the spatiotemporal correlations of adjacent regions:
\begin{equation}\label{eq:NTA-1}
\begin{gathered}
    \widetilde Q^{(l),h}=W^{(l),\widetilde Q}_hF_{PE}(\widetilde M^{(l),t}_{r_i}),\\
    \widetilde K^{(l),h}=W^{(l),\widetilde K}_hF_{PE}(\widetilde M^{(l),t'}_{r_j}),\\
    \widetilde V^{(l),h}=W^{(l),\widetilde V}_hF_{PE}(\widetilde M^{(l),t'}_{r_j}),\\
\end{gathered}
\end{equation}
where $W^{(l),*}_h\in\mathbb{R}^{d/H\times d}$ is the parameters in the $l$-th layer and the $h$-th head. $\widetilde M^{(l),t}_{r_i}$ denotes the output of the time-aware recurrent module of the $r_i$-th region in the $t$-th time slot. $F_{PE}$ denotes the positional encoding operation. $\widetilde M^{(l),t'}_{r_j}$ represents the potential information of the $r_j$-th region in the $t'$-th time slot. By this design, we then calculate the correlation between the 
region $r_i$ and its neighbors at each time interval. Similar to Eq.~\ref{eq:MTC-2}, the output are calculated as: 
\begin{equation} \label{eq:NTA-2}
\begin{gathered}
    \hat{score}^{h}_{t,t'}=\frac{\widetilde Q^{(l),h\top }\widetilde {K}^{(l),h}}{\sqrt{d/H}},\\
\hat{\alpha}^{h}_{t,t'}=softmax(\hat{score}^{h}_{t,t'})=\frac{exp(\hat{score}^{h}_{t,t'})}{\sum_{t'}exp(\hat{score}^{h}_{t,t'})},\\
    \hat{m}^{h}_{t}=\sum_{t'=1}^T\hat{\alpha}^{h}_{t,t'}\cdot \widetilde V^{(l),h},
\end{gathered}
\end{equation}
where $\hat{\alpha}^{h}_{t,t'}$ is the attention weight between the $t$-th and $t'$-th time slot in the $h$-th head. We then get the multi-head representation by the following formula:
\begin{equation}\label{eq:NTA-3}
    \hat{M}^{(l)}_t={\big\Arrowvert}^H_{h=1}m^{(l),h}_{t}\cdot W^{(l),\widetilde O}
\end{equation}
where $\big\Arrowvert$ represents the concatenation operation, $W^{(l),\widetilde O}\in\mathbb{R}^{d\times d}$ is the parameter matrices. 

\par Finally, for the target region $r_i$, we generate a weighted aggregation of the feature representations of its neighbors through the attention mechanism, where the influence weight $\widetilde\alpha$ is calculated by Eq.~\ref{eq:GAT-1}. Formally, we have the following formula:
\begin{equation}\label{eq:GAT-2}
    E^{(l)}_{r_i}=\sigma\left(\sum_{r_k\in \mathcal{N}_{r_i}}\widetilde\alpha_{r_i,r_k}\hat{M}^{(l)}_{r_k}\right)
\end{equation}
where $\widetilde\alpha_{r_i,r_k}$ is the influence weight of neighbor $r_k$ to $r_i$. $\mathcal{N}_{r_i}$ is the set of $r_i$'s neighbor nodes. $\hat{M}^{(l)}_{r_k}$ is the urban anomaly embeddings associated with region $r_k$ generated by Eq.~\ref{eq:NTA-3}. We denote $E^{(l)}=\functhree(\widetilde M^{(l)})$, i.e., Eq.~\ref{eq:GAT-1}-\ref{eq:GAT-2} as the function $\functhree$ mentioned in Eq.\ref{eq:layerC}.
\par After $L$ iterations according to Eq.\ref{eq:layerC}, we get the final output $E^{(0)},E^{(1)}, \cdots,E^{(L)}\in\mathbb{R}^{N\times T\times C \times d}$ of the STC module.

\subsection{Multi-Task Prediction}

\par To solve the zero-inflated issue, we design a multi-task module during prediction inspired by zero-inflated Poisson (ZIP) regression~\cite{lichman2018prediction}, which is used to predict the user-item rate. 
Similarly, the number of examples with $urban\ anomaly\ index=0$ dominate in the real-world urban anomaly datasets. In other words, the urban anomaly datasets are quite sparse, hence using a single model to simulate the whole datasets is difficult. 

Therefore, we design a multi-task framework to simulate the mixture of the two processes in urban anomaly prediction: an exposure process and a count process. We use the count process as the auxiliary task to predict whether urban anomaly would occur, and the exposure process as the main task to predict the value of urban anomaly index.

\textbf{Exposure Process.} The exposure process describes whether or not a region $r$ has been exposed to the $c$-th type at the $t$-th time slot. It is found on the premise that, because urban anomaly is uncommon, the vast majority of the regions lack the conditions that would allow urban anomaly to occur at a specific time, such as the existence of potential criminals, weak law and order, and so on. That is, a certain type $c$ is not exposed to the region $r$ in the current time slot $t$, and it is classified as no urban anomaly in terms of the binary classification problem. Here, we define $z^t_{r,c}\in\{0,1\}$ to denote whether a type $c$ is exposed to a region $r$ at time $t$.

\textbf{Count Process.} Conditioned on exposure, i.e., $z^t_{r,c}=1$, count process then accounts for the specific number of the $c$-th type at the $r$-th region in the $t$-th time slot. That is, we use the regression task (main task) to forecast the urban anomaly indices when the binary classifier indicates that urban anomaly will occur (auxiliary task).

\textbf{Multi-Task Prediction.}
Based on the main and auxiliary tasks, our multi-task prediction framework works as follows.
Firstly, we feed the output feature maps of the STC module including $(E^{(0)}, \cdots,E^{(L)})$ into the multi-task prediction module as shown in Figure~\ref{fig:model}. Firstly, we average the embeddings $(E^{(0)}, \cdots,E^{(L)})$. Then, we sum up the result along the time dimension to generate the final embeddings $E^{T+1}\in\mathbb{R}^{N\times C\times d}$.

Secondly, with the auxiliary task, we make predictions of different types of a urban anomaly  as
\begin{equation}
    P^{T+1}_c=\sigma(W_{c}^\top E^{T+1}_c),
    \nonumber
\end{equation}
where $\sigma$ denotes the sigmoid function, and $p^t_{r,c}$ represents the probability of $z^t_{r,c}=1$. 
We can convert $P$ to $Z$ using the hyperparameter threshold. 

Thirdly, with the main task, we predict the number of different types of a urban anomaly as
\begin{equation}
    \hat{X}^{T+1}_c=W_c^\top Z_c E^{T+1}_c,
    \nonumber
\end{equation}
where the multiplication with the $Z_c$ matrix is to mask the combinations of time slots and regions that do not have criminal conditions. 

Lastly, we divide the data distribution into four subsets $F$ according to the urban anomaly indices and then set the weights on the samples accordingly. With these weights, we define the loss function of the main task with the goal of further reducing the impact of excessive zero data on the parameter learning. 
More specifically, the first three subsets correspond to the cases where the  urban anomaly index are 0, 1, and 2, respectively; and the last subset corresponds to the urban anomaly index equals to or is greater than 3. We denote the four subsets by $F=\{0,1,2,\geq3\}$. 
Then, \sysname is optimized by minimizing the total loss of the multi-task target:
\begin{equation}
    \begin{gathered}
        L_c=-\sum_t\delta(X^t)logP^t+\bar\delta(X^t)log(1-P^t),\\
        L_r=\sum_{f\in F}\left (\eta_f\sum_t\big\Arrowvert X^t_f-\hat{X}^t_f\big\Arrowvert^2_2\right),\\
        Loss=L_r+\lambda_cL_c+\lambda_{reg}\big\Arrowvert\Theta\big\Arrowvert^2_2,\\
    \end{gathered}
\end{equation}
where $X^t$ denotes the ground truth, $P^t$ and $\hat{X}^t$ denote the predicted results of auxiliary and main tasks in the $t$-th time slot, respectively. The symbols with the subscript $f$ indicate those samples whose urban anomaly index is $f\in F$. $\eta_{f\in F}$ indicate the weight. $\delta$ and $\bar\delta$ represent element-wise positive and negative indicator functions, respectively. $L_c$ and $L_r$ are loss functions of the auxiliary task and main task, respectively. The final loss function to minimize is $Loss$. We use L2 regularization, i.e., $\big\Arrowvert\Theta\big\Arrowvert^2_2$ to avoid overfitting. $\lambda_c$ and $\lambda_{reg}$ are the hyperparameters of the loss function which are the weights of auxiliary task loss $L_c$ and L2 regularization, respectively.

\section{EXPERIMENTS ON CRIME PREDICTION}\label{sec:crime}

In this section, we conduct extensive experiments on two real-world crime datasets collected from New York City and Chicago, and analyze the results to validate the effectiveness of \sysname. We perform experiments to answer the following questions:
\begin{itemize}
    \item \textbf{Q1}: Compared to traditional methods and SOTA technologies, can \sysname be more effective in solving the zero-inflated problem and achieve comparable performance?
    \item \textbf{Q2}: How do different components affect the forecasting performance of \sysname?
    \item \textbf{Q3}: How do different settings of hyperparameters affect the performance of \sysname?
    \item \textbf{Q4}: Can \sysname provide the model interpretability w.r.t spatial, temporal and semantic dimensions?
\end{itemize}

\subsection{Experimental Setup}\label{sec:setup}

\noindent \textbf{Datasets.} 
We evaluate \sysname on two real-world crime datasets collected from Jan 1, 2014 to Dec 30, 2015 in New York City~\cite{xia21spatial} and Chicago~\footnote{https://data.cityofchicago.org/}, referred as NYC and CHI, respectively. In our experiments, we focus on several key categories and consider four categories of crimes in each dataset. 

In the NYC dataset, New York City is partitioned into $N=127$ disjointed geographical regions using the grid-based mapping approach as described in Section~\ref{sec:DP}, and each region is a $3km\times 3km$ spatial unit. We choose four categories of crimes that are interconnected and clearly affected by spatiotemporal factors: burglary, robbery, assault, and larceny. 

The CHI dataset is widely applied in a variety of crime prediction tasks. According to the administrative division plan enacted by the local government, that is, the region-based mapping approach described in Section~\ref{sec:DP}, Chicago is divided into $N=77$ disjointed geographical regions. In the same way as NYC, we choose four categories of crime: burglary, narcotics, assault, and robbery. Descriptive statistics of the two datasets are summarized in Table~\ref{tab:dataset}.

\begin{table}[!t]
\renewcommand{\arraystretch}{1.3}
  \caption{Statistics of Crime Datasets.}
  \label{tab:dataset}
  \centering
  \begin{tabular}{ccccc}
    \toprule
    Data&\multicolumn{2}{c}{NYC-Crime Reports}&\multicolumn{2}{c}{CHI-Crime Reports}\\
    \midrule
    Time Span&\multicolumn{2}{c}{Jan, 2014 to Dec, 2015}&\multicolumn{2}{c}{Jan, 2014 to Dec, 2015}\\
    Regions \#/Type&\multicolumn{2}{c}{256($16\times16$)/Grid}&\multicolumn{2}{c}{77/Region}\\
    Category \#&\multicolumn{2}{c}{4}&\multicolumn{2}{c}{4}\\
    Category&Burglary&Robbery&Burglary&Narcotics\\
    Instances \#&31,799&33,453&33,807&27,610\\
    Category&Assault&Larceny&Assault&Robbery\\
    Instances \#&40,429&85,899&28,888&19,367\\
  \bottomrule
\end{tabular}
\end{table}

\noindent\textbf{\sysname Implementation.} 
We implement \sysname using the TensorFlow framework and train our model with 
the Adam optimizer. The learning rate is initialized to 0.001 with a decay rate 0.96. 

We run experiments on a server equipped with an Nvidia GeForce RTX 3090 GPU.
We use one day as the time slot and use a month's data to predict the number of crimes in the following day. That is, the parameter $T$ is set to 30. We partition all data on the time axis into the training and test sets by a ratio of 7:1, and use the last 30 days of the training set for validation. 
All data is normalized into the range $[0, 1]$ by z-score normalization. 
The default hyperparameters settings of \sysname are as follows: the size of hidden state $d$ is set as 16. In the STC module, we iterated the STC layer 3 times, i.e., $L=3$. In all multi-head self-attention mechanisms, the number of heads $H = 8$. 
In the loss function, the weight decay factors $\lambda_*$ are searched in the range of \{0, 0.0001, 0.001, 0.01\}. According to ~\cite{wang2021gsnet}, the weights of different samples in the loss function of the main task $\eta_{f\in F}$ are set to 0.05, 0.2, 0.25 and 0.5, respectively.

\vskip 0.1in\noindent \textbf{Baselines.} 
To demonstrate the effectiveness of our model, we measure \sysname against multiple traditional methods and SOTA methods from five categories: (i) conventional time-series prediction technique (SVR); (ii) conventional deep learning method (MLP); (iii) conventional recurrent neural network technique (GRU); (iv) spatial-temporal prediction methods with GNNs (GMAN and ST-SHN); (v) spatial-temporal prediction method considering the zero-inflated issue (MAPSED). The details are as follows:
\begin{enumerate}
    \item Support Vector Regression (SVR) is a classic machine learning algorithm, especially for time-series forecasting by transforming data into feature space.
    \item Multiple Layer Perception (MLP) is the fundamental method of DNN, which comprises three layers: input, hidden, and output, all of which are fully connected, and the layers are fully connected. We select ReLU as its activation function. 
    \item Gated Recurrent Unit (GRU) is a typical RNN framework that is frequently used for time-series forecasting. It has a superior long-term memory than the original RNN and is more efficient at solving the gradient issue.
    \item Graph Multi-Attention Network (GMAN)~\cite{zheng2020gman} adapts an encoder-decoder architecture based on the graph multiple spatiotemporal attention blocks to model the impact of the spatiotemporal factors.
    \item Spatial-Temporal Sequential Hypergraph Network( ST-SHN)~\cite{xia21spatial} is a crime prediction framework with the hypergraph learning paradigm to enhance cross-region relation learning without the limitation of adjacent connections and the multi-channel routing mechanism to learn the time-evolving structural dependency across crime types.
    \item Multi-axis Attentive Prediction for Sparse Event Data (MAPSED)~\cite{DBLP:journals/corr/abs-2110-01794} achieves strong generalization in the sparse observations by applying a tensor-centric, fully attentional approach to extract the short-term dynamics and long-term semantics of event propagation.
\end{enumerate}

\vskip 0.1in\noindent\textbf{Evaluation Metrics.} 
To evaluate the performance in predicting the quantitative number of crimes, we utilize two metrics, i.e., MAE and RMSE. MAE is a general form for measuring the mean error, whereas RMSE penalizes unusual samples more. 
For both metrics, lower values suggest better performance. We briefly describe the metrics as follows:
\begin{equation} 
\begin{gathered}
    MAE=\frac{1}{n}\sum_i^n\left|X_i-\hat{X}_i\right|,\\
    RMSE=\sqrt{\frac{1}{n}\sum_i^n\left(X_i-\hat{X}_i\right)^2},\\
\end{gathered}
\end{equation}
where $n$ denotes the number of all samples, and $X_i$ and $\hat{X}_i$ denote the ground truth and the predicted value of the $i$-th sample, respectively.

\begin{table*}[ht]
\renewcommand{\arraystretch}{1.3}
\caption{Performance comparison of \sysname and baselines for crime forecasting. RMSE* and MAE* represent the performance on non-zero samples, while RMSE and MAE represent the performance with zero inflation.}
\label{tab:performance}
\centering
\begin{tabular}{ccccccccc}
\toprule
\multicolumn{2}{c}{Method}    & \multirow{2}{*}{SVR} & \multirow{2}{*}{MLP} & \multirow{2}{*}{GRU} & \multirow{2}{*}{GMAN} & \multirow{2}{*}{ST-SHN} & \multirow{2}{*}{MAPSED} & \multirow{2}{*}{\sysname} \\
Dataset              & Metric &                      &                      &                      &                       &                         &                         &                       \\
\cmidrule(r){1-9}
\multirow{4}{*}{NYC} & RMSE   & 1.1264               & 1.3558               & 1.0762               & 0.9685                & 0.9754                  & 0.9061                  & \textbf{0.8632}       \\
                     & MAE    & 0.7948               & 0.9147               & 0.7836               & 0.6821                & 0.6807                  & 0.6054                  & \textbf{0.5655}       \\
\cmidrule(r){2-9}
                     & RMSE*  & 1.7221               & 2.5175               & 0.9585               & 0.9450                & 0.8997                  & 0.9112                  & \textbf{0.8591}       \\
                     & MAE*   & 1.1311               & 1.3020               & 0.6706               & 0.6645                & 0.5867                  & 0.6340                  & \textbf{0.5824}       \\
\cmidrule(r){1-9}
\multirow{4}{*}{CHI} & RMSE   & 0.9152               & 0.9936               & 1.0479               & 0.9432                & 0.9997                  & 0.8774                  & \textbf{0.7826}       \\
                     & MAE    & 0.7097               & 0.8009               & 0.7687               & 0.6732                & 0.7294                  & 0.5915                  & \textbf{0.5122}       \\
\cmidrule(r){2-9}
                     & RMSE*  & 1.1047               & 1.1183               & 0.8238               & 0.8123                & 0.7920                  & 0.8943                  & \textbf{0.7904}       \\
                     & MAE*   & 0.7936               & 0.7932               & 0.6019               & 0.6012                & 0.5395                  & 0.5953                  & \textbf{0.5166}   \\   
\bottomrule
\end{tabular}
\end{table*}

We notice that, trained with both zero-inflated dataset and common dataset, some baselines tend to make close-to-zero prediction thus perform better on zero-inflated dataset while some baselines perform well on common dataset yet cannot effectively learn zero-inflated data. To better understand the performance difference, we execute comparative experiments with two derived metrics MAE* and RMSE*, where * indicates the performance on non-zero samples. 

The RMSE and MAE of the ideal prediction model should be as small as possible, indicating that the model performs well on the entire zero-inflated data. At the same time, its RMSE* and MAE* should not be too large, indicating that the model can correctly predict non-zero values without affected by too many zero-value samples.

\begin{figure}[t!]
    \centering
    \includegraphics[width=0.85\linewidth]{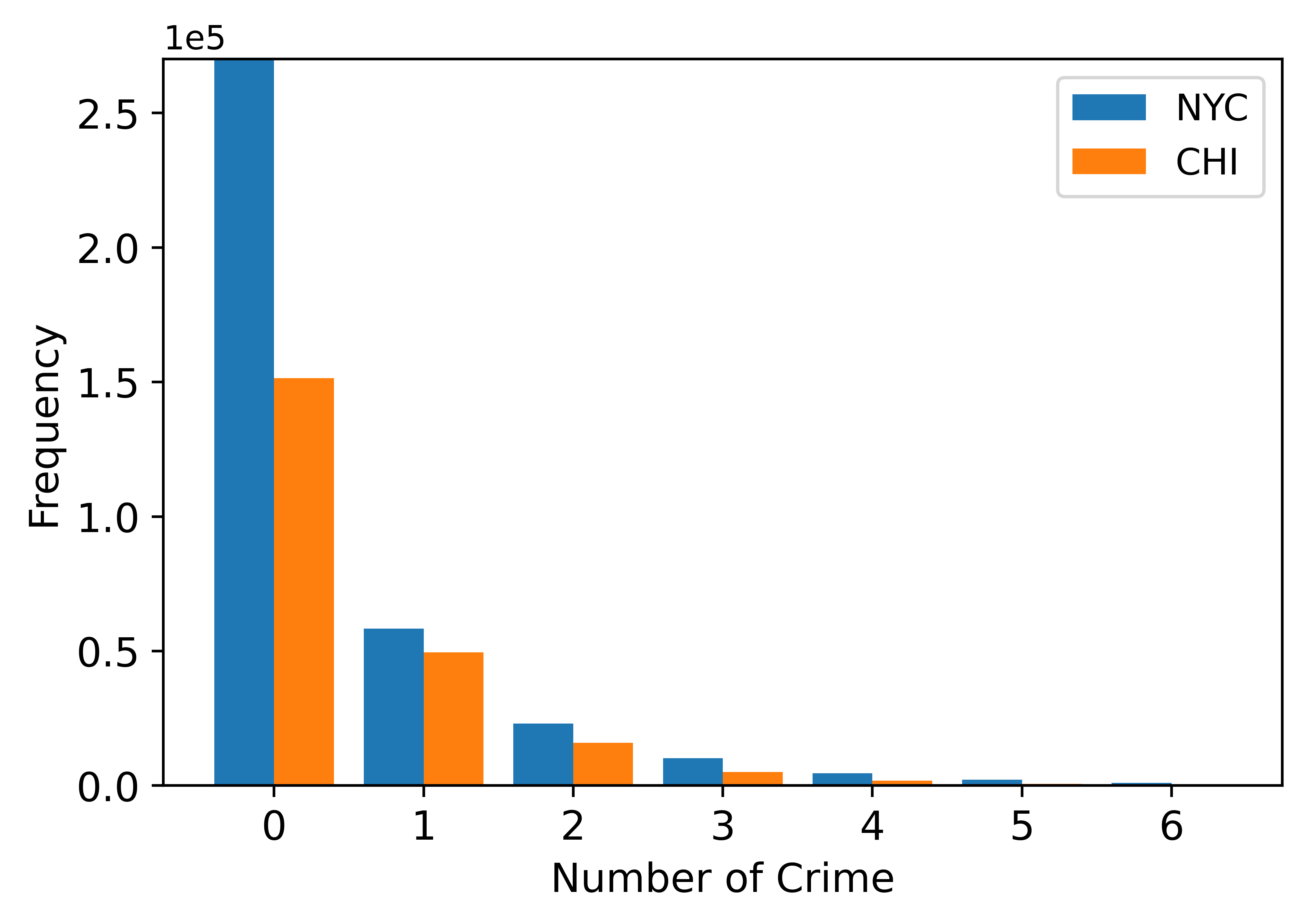}
    \vspace{-0.1in}
    \caption{The zero-inflated issue in two real-world crime datasets collected from Jan 1, 2014 to Dec 30, 2015 in New York City (NYC)~\cite{xia21spatial} and Chicago (CHI).}
    \label{fig:zero-inflated}
    \vspace{-0.2in}
\end{figure}

\subsection{Performance Comparison for Crime Forecasting (Q1)}

We quantify the characteristics of zero inflation in the data sets, as zero inflation has significant impacts on model performance. 
Figure~\ref{fig:zero-inflated} shows the distribution of crime occurrences. Both datasets meet the characteristics of zero-inflated data whose ratios of zero data are 72.7\% and 67.4\%, respectively. Because of its finer granularity of spatial division, the NYC dataset has a larger percentage of zero data.

\begin{figure}[t!]  
    \centering    
    \includegraphics[width=\linewidth]{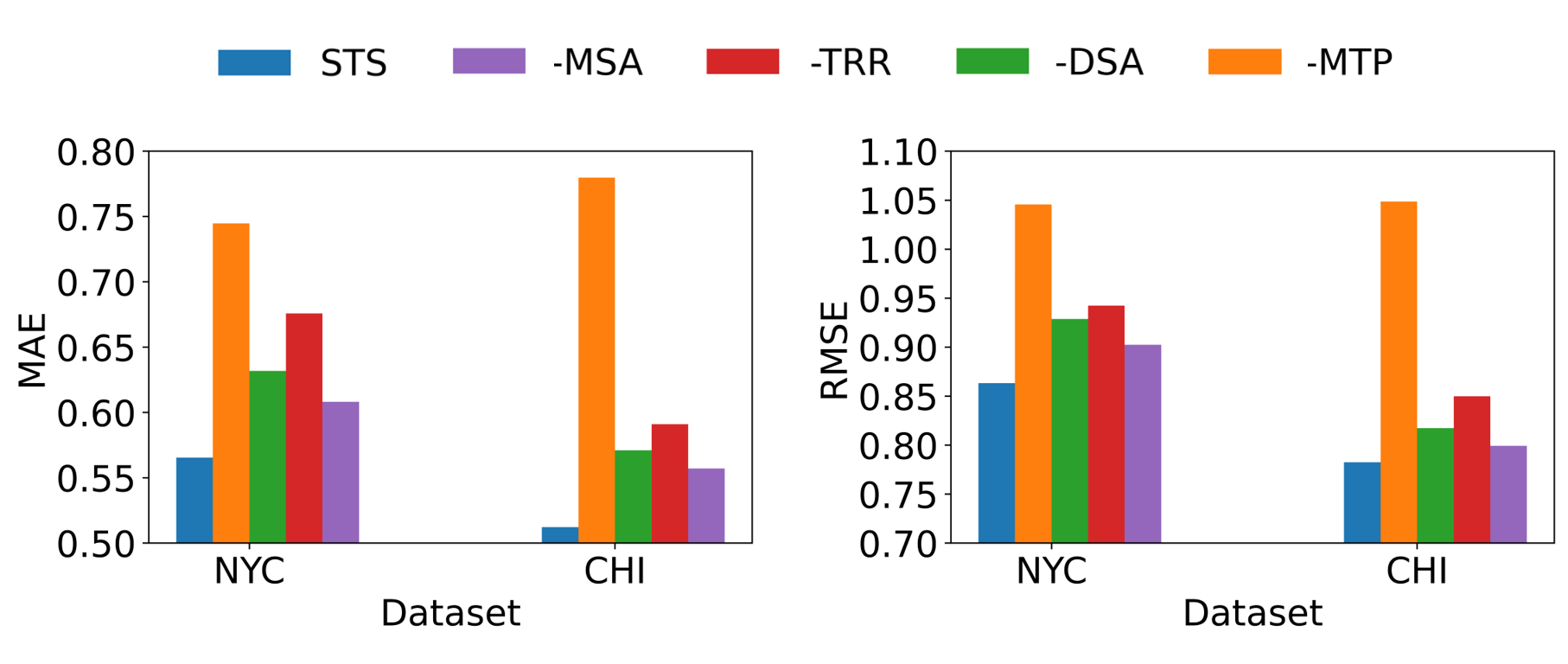}
    \vspace{-0.25in}
    \caption{Model Ablation Study.}
    \label{fig:ablation} 
    \vspace{-0.2in}
\end{figure}

\begin{figure*}[!t]
\centering
\subfloat[Number of STC Layers]{\includegraphics[width=0.3\textwidth]{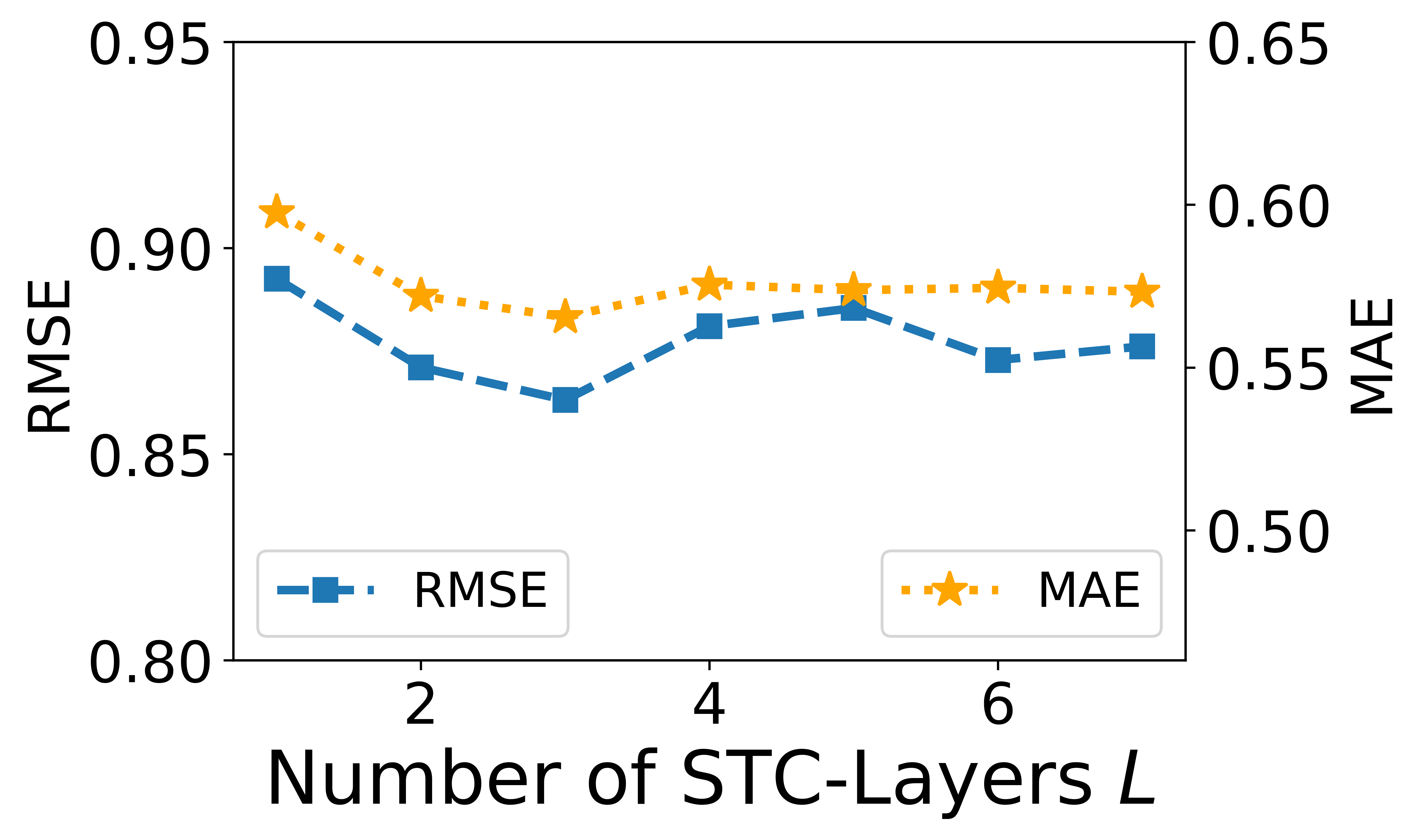}}
\hfil
\subfloat[Number of Head]{\includegraphics[width=0.3\textwidth]{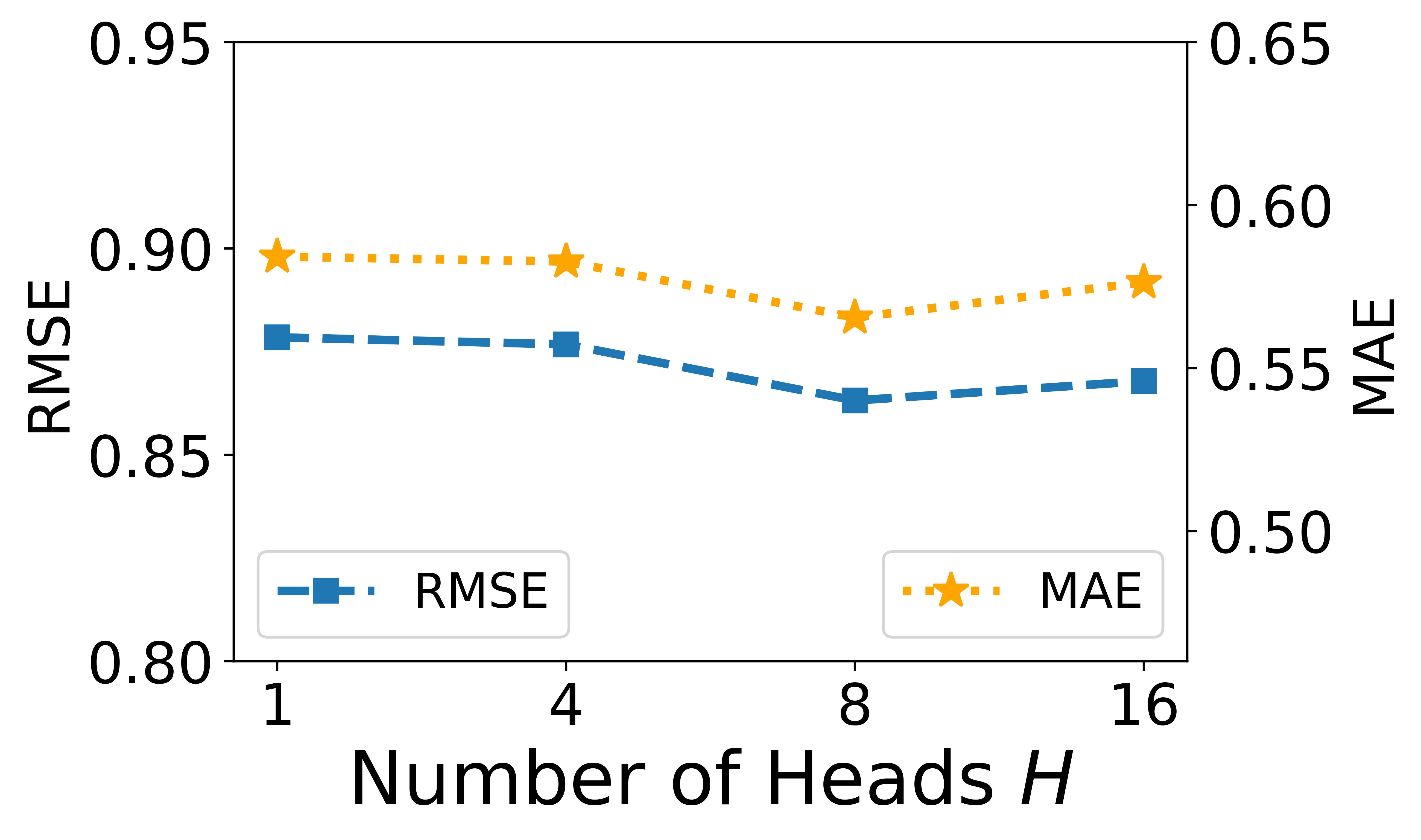}}
\hfil
\subfloat[Hidden State Dimension]{\includegraphics[width=0.3\textwidth]{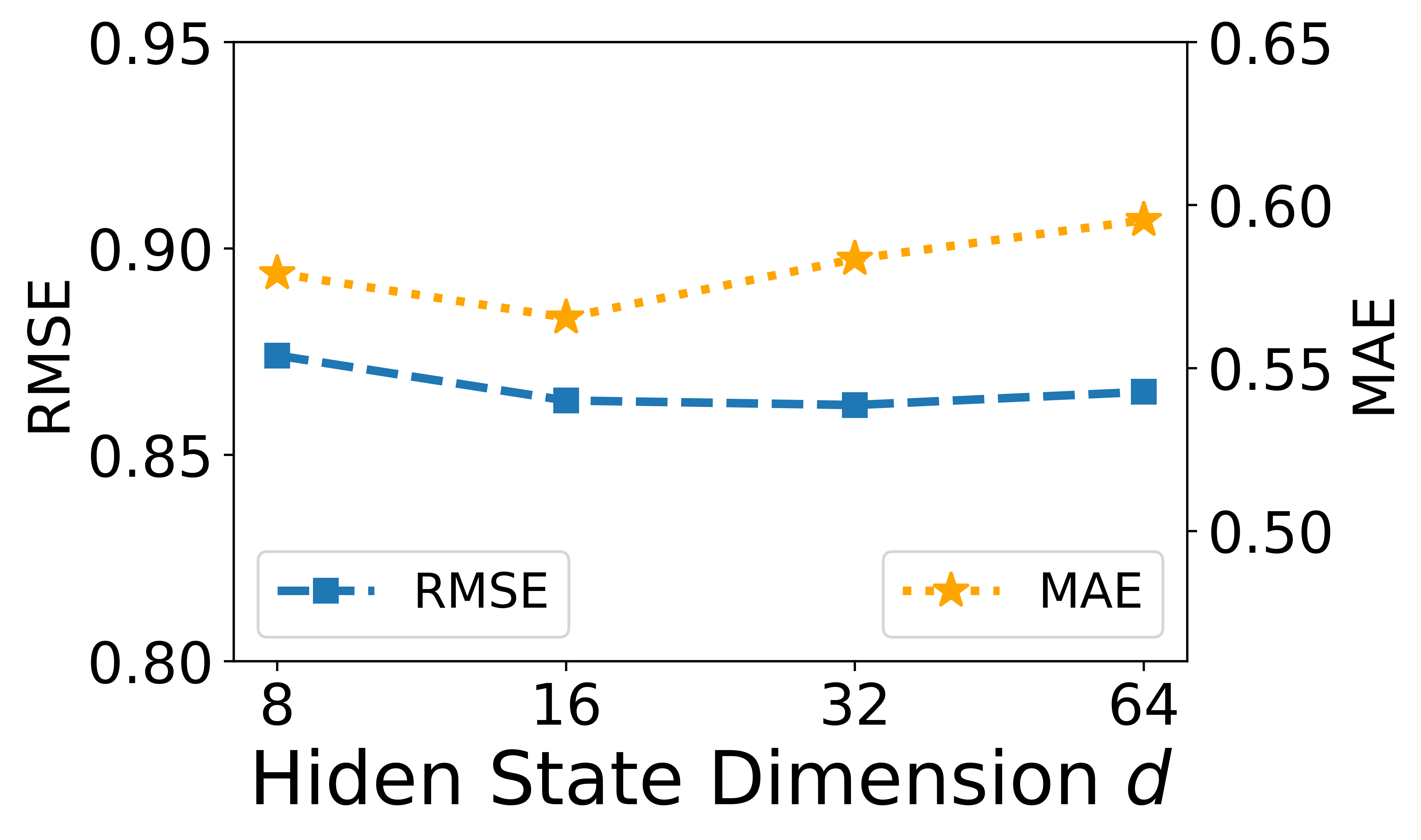}}
\caption{Hyperparameters studies on NYC. Performance on CHI is similar.}
\label{fig:parameterCrime}
\end{figure*}

We next conduct experiments and summarize the performance results in Table~\ref{tab:performance}. 
We make the following observations.

\textbf{First}, we observe that \sysname significantly outperforms all baseline models on both datasets and on both metrics. More specifically, \sysname improves the MAE metric by 26.02\% and the RMSE metric by 18.95\%, respectively. This clearly demonstrates the advantages of our model and suggests that our design for \sysname is effective.

\textbf{Second}, we observe that zero inflation does have significant negative impacts on the performance of all baseline models: the performance of RMSE and MAE on all samples is consistently much worse than that on non-zero samples in the methods that do not consider the zero-inflated issue. This result illustrates that these methods cannot actually learn crime patterns well and demonstrates the importance of solving the zero-inflated issue.

More specifically, the traditional statistical method SVR and the simple fully connected neural network MLP suffer more from zero inflation and show the largest gap between the two kinds of samples. 
For instance, on the NYC dataset, the performance of SVR and MLP are improved by approximately 56\% when zero inflation is removed, suggesting that zero inflation can significantly degrade their performance.
On the CHI dataset, SVR and MLP perform even better than some deep learning baselines on entire samples. In fact, they simply predict the number of crimes to be all zeros after flooring~\cite{DBLP:journals/corr/abs-2110-01794}, thus achieving the relatively low RMSE and MAE. Then, GRU can capture the temporal dependence of time-series, GMAN models temporal and spatial dependence respectively, and ST-SHN extracts not only spatiotemporal dependence but also semantic correlation. Thus their ability to model crime patterns is gradually enhanced and their performance is improved. However, they still perform poorly on the zero-inflated dataset.

\textbf{Third}, MAPSED is an exception in that it performs particularly well on zero inflated samples. However, it only uses deep learning methods in the model that are experimentally more suitable for zero-inflated data, but does not design a dedicated module to solve this issue, which explains its low performance compared with \sysname.
Additionally, MLP performs the worst overall due to the fact that it is not applicable to spatiotemporal data.

\subsection{Model Ablation Study (Q2)}

We consider four different variants to evaluate the effectiveness of the components in \sysname: (i)-MSA: we remove the multi-head semantic self-attention which capture the intrinsic links between different crime types; (ii)-TRR: we remove the time-aware recurrent module which is used to capture the short-term time dependence; (iii)-DSA: we remove the dynamic spatiotemporal attention in the STC module; (iv)-MTP: instead of the multi-task prediction module which is used to handle the zero-inflated issue, we simply apply a linear regression to predict the result.

We summarize in Figure~\ref{fig:ablation} the results of model ablation study.
We observe that the integrated \sysname achieves the best performance. 
More specifically, the replacement of the multi-task prediction module has the largest impact on the performance, which proves that the proposed multi-task prediction module effectively solves the zero-inflated problem in crime forecasting. Removing the DSA, TRR, and MSA components also impact the performance of \sysname, which demonstrates that simultaneously capturing the temporal, geospatial, and semantic correlations is critical to crime prediction. Among the three variants, the -TRR variant has the worst performance, suggesting that short-term temporal correlation is more relevant with the crime prediction task.

\subsection{Hyperparameters Studies (Q3)}

We design experiments to investigate the impacts of hyperparameter tuning on the prediction performance of \sysname and verify its robustness. When tuning one of the hyperparameters, the others remain the default values.
We summarize the results of the hyperparameters studies on NYC in Figure~\ref{fig:parameterCrime} and make the following observations.

First, \sysname does not show considerable performance fluctuation during the tuning, which suggests that \sysname is robust.
%
Second, when we gradually vary the number of the STC layers $L$ from 1 to 7, the performance first rises and then falls, with the best performance at $L=3$. This is because with too few layers, the model can only aggregate the influence of neighbors within a few number of hops, thus lacking the strong ability to capture sufficient geospatial correlations. On the other hand, with too many layers, the model absorbs too much noise learned from distant neighbors and thus its performance is degraded. 
Note that a model with too many layers needs to train a much larger number of parameters and is thus prone to become over-fitted. Similarly, when we vary the number of heads in attention mechanism $H$ and the hidden state dimension $d$, it also appears that the performance of \sysname first gets better and then falls back. The reason is, with too few heads or small dimension, the model is under-expressed, while on the contrary the model becomes over-fitted. The best results of \sysname are achieved when $H=8$ and $d=16$, respectively. 
The performance results on CHI are similar and we omit the results due to space limit.

\begin{figure*}[!t] 
    \centering
    \includegraphics[width=\linewidth]{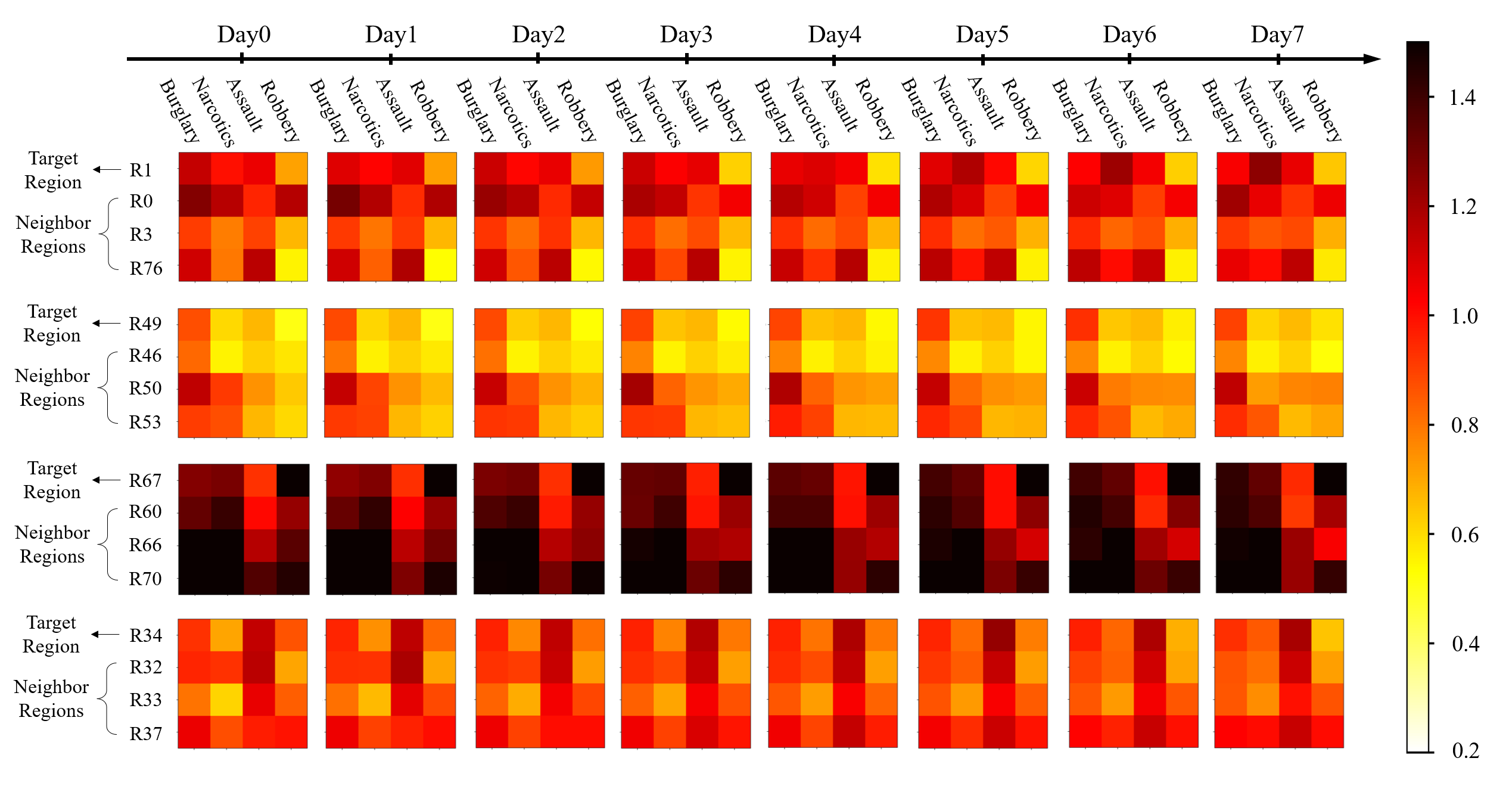}
    \vspace{-0.2in}
    \caption{Visualization for dependencies captured by \sysname in spatial, temporal, and semantic dimensions. Squares represent the occurrence of a type of crimes predicted by \sysname in a region on a given day.}
    \label{fig:heatmap}
    \vspace{-0.1in}
\end{figure*}

\begin{figure}[!t] 
    \centering
    \includegraphics[width=0.95\linewidth]{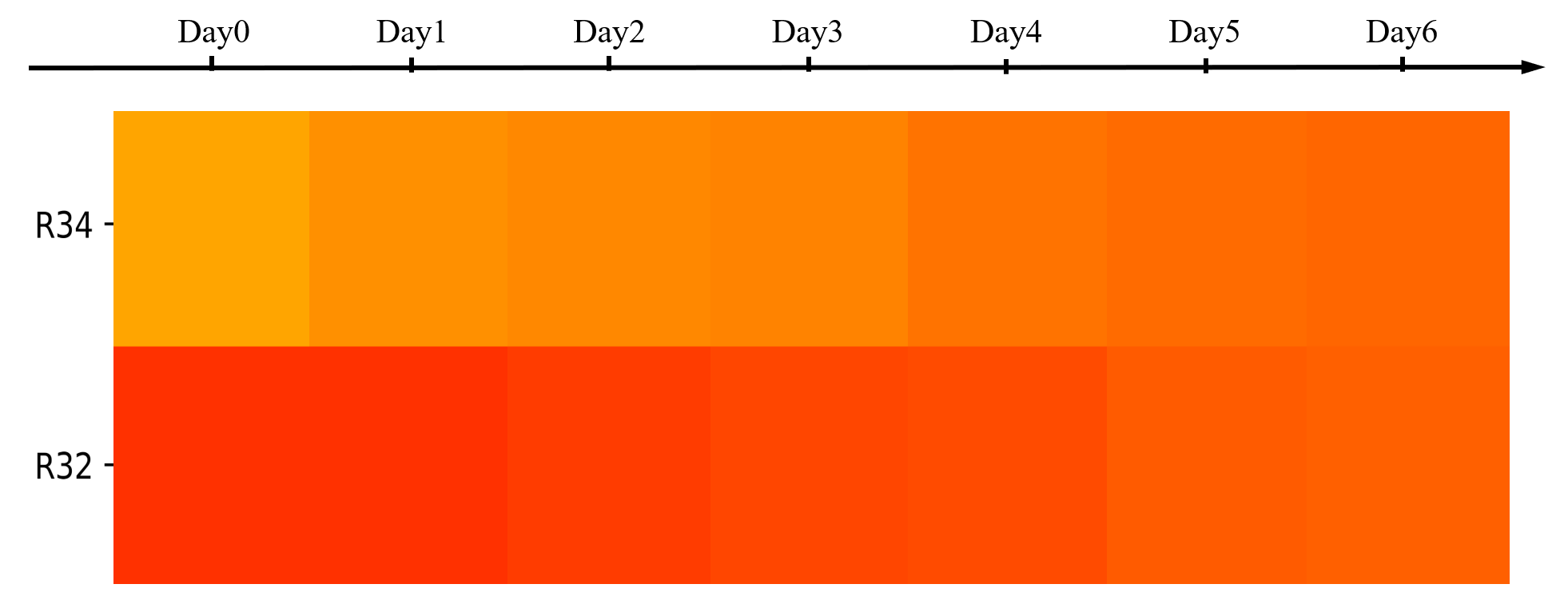}
    \renewcommand{\arraystretch}{1.3}
    \caption{Spatiotemporal examples.}
    \label{fig:spati-temporal examples}
\end{figure}

\subsection{Interpretability (Q4)}

\par Previous experiments have verified that \sysname achieves superior performance, but does \sysname indeed jointly capture the correlations in the spatial, temporal, and semantic dimensions? We offer several examples in CHI dataset to give an intuitive impression to illustrate the interpretability of \sysname.

Figure~\ref{fig:heatmap} shows the interpretability of \sysname employing a heatmap taking four target regions with different crime densities (R1, R49, R67, R34) as examples. Each set of samples contains eight $4\times4$ subgraphs, corresponding to eight consecutive time slots. In each subgraph, the first row represents the target region, and the last three rows correspond to three neighboring regions of the target region. In each subgraph, the four columns indicate the four crime categories selected in CHI. Darker color in a square indicates a larger number of a certain type of crimes in a region predicted by \sysname in a certain time slot. 

We make the following observations.
Firstly, we observe that each small square in the $4\times4$ subgraph of a selected sample has a smooth color change over time, indicating that \sysname sufficiently capture the correlation in the temporal dimension. 

Secondly, in the spatial dimension, the target region exhibits a more similar color with its neighbor in the next time slot because of the inter-region affect. An example of the narcotics type is shown in Figure~\ref{fig:spati-temporal examples}. The predicted number of crimes in region R34 is small at $day0$ and gradually increases in subsequent time slots due to the influence of neighbor R32. The predicted value of R32 also gradually decreases in subsequent time slots due to the influence of R34. This indicates that \sysname dynamically captures the spatial correlation as time advances. 

Thirdly, in the semantic dimension, the prediction results that absorbed the correlations across different crime types become similar over time. This demonstrates that \sysname also dynamically captures semantic dependencies. Finally, 
the color of each square changes under the joint influence of the three dimensions. 
As such, \sysname can jointly capture correlations from these three dimensions.

\subsection{Summary}

The evaluation results validate the effectiveness of \sysname on two real-world zero-inflated crime datasets. Specifically, \sysname improves on average by 26.02\% and 18.95\% over baselines in MAE and RMSE metrics, respectively. 
We observe that all components of \sysname bring gains, with the multi-task prediction module bringing the biggest boost. We then experimentally demonstrate the robustness of \sysname and analyze the effect of different settings of hyperparameters on the model. Finally, we show that \sysname can provide interpretability in terms of spatial, temporal, and semantic dimensions.

\section{EXPERIMENTS ON TRAFFIC ACCIDENT RISK PREDICTION}\label{sec:traffic}
In this section, we conduct extensive experiments on two public real-world traffic accident risk datasets collected from New York City and Chicago, and analyze the results to validate the effectiveness of \sysname. We skip the overlapped parts with the previous experiment. 

\subsection{Experimental Setup}

\noindent \textbf{Datasets.} 
We evaluate \sysname on two real-world traffic accident risk datasets collected from New York City~\footnote{https://opendata.cityofnewyork.us/} and Chicago~\footnote{https://data.cityofchicago.org/}, referred as NYC and CHI, respectively. The traffic accident data includes location, time and the number of casualties. In the experiments on traffic accident risk, we remove the module used to extract semantic correlations between different categories, i.e., the MSA module, because the data do not distinguish between categories.

According to the number of casualties in traffic accidents, we define three traffic accident types, i.e., minor accidents, injured accidents and fatal accidents, and corresponding risk values are set to be 1, 2 and 3 respectively. In the task of traffic accident risk prediction, the urban anomaly indices $X^t_r$ is the sum of traffic accident risk values in region $r$ at time interval $t$.
For example, $X^t_r$ is 5 if two minor accidents and one fatal
accident have happened in region $r$ at time interval $t$.~\cite{wang2021gsnet}

In the NYC dataset, New York City is partitioned into $N=243$ disjointed geographical regions using the region-based mapping approach as described in Section~\ref{sec:DP}, and each region is a $2km\times 2km$ spatial unit.

In the CHI dataset, according to the region-based mapping approach described in Section~\ref{sec:DP}, Chicago is split into $N=197$ rectangle regions. In the same way as NYC, each region is a $2km\times 2km$ spatial unit. Descriptive statistics of the two datasets are summarized in Table~\ref{tab:trafficDataset}.

\begin{table}[t!]
  \caption{Statistics of traffic accident risk datasets.}
  \label{tab:trafficDataset}
  \centering
  \begin{tabular}{ccc}
    \toprule
    Data & \makecell[c]{NYC-Traffic \\ Accident Reports} & \makecell[c]{CHI-Traffic \\ Accident Reports}\\
    \midrule
    Time Span&Jan, 2013 to Dec, 2013&Feb, 2016 to Sept, 2016\\
    Road Segment \#&103k&56k\\
    Regions \#&243&197\\
    Instances \#&147k&44k\\
    Category \#&1&1\\
  \bottomrule
\end{tabular}
\end{table}

\noindent \textbf{\sysname Implementation.} 
We implement \sysname using the TensorFlow framework and train our model with 
the Adam optimizer. The learning rate is initialized to 1e-5 with a decay rate 0.96. 

We run experiments on a server equipped with an Nvidia GeForce RTX 3090 GPU.
We use 6 hours as the time slot and use 30 time slots of data to predict the value of traffic acciednt risk in the following one time slot. That is, the parameter $T$ is set to 30. We partition all data on the time axis into the training and test sets by a ratio of 7:1, and use the last 30 time slots of the training set for validation. 
All data is normalized into the range $[0, 1]$ by Max-Min normalization. 
The default hyperparameters settings of \sysname are as follows: the size of hidden state $d$ is set as 16. In the STC module, we iterated the STC layer 2 times, i.e., $L=2$. In all multi-head self-attention mechanisms, the number of heads $H = 4$. 
In the loss function, the weight decay factors $\lambda_*$ are searched in the range of \{0, 0.0001, 0.001, 0.01\}. According to ~\cite{wang2021gsnet}, the weights of different samples in the loss function of the main task $\eta_{f\in F}$ are set to 0.05, 0.2, 0.25 and 0.5, respectively.

\vskip 0.1in\noindent\textbf{Baselines.} We use the same baselines as in the previous experiment.
%

\vskip 0.1in\noindent\textbf{Evaluation Metrics.} 
To evaluate the performance in predicting the quantitative values of the traffic accident risk, we utilize two metrics, i.e., MAE and RMSE. Similar to the crime prediction task, we execute comparative experiments with two derived metrics MAE* and RMSE*, where * indicates the performance on non-zero samples. 

\subsection{Performance Comparison for Traffic Accident Risk Forecasting (Q1)}

We quantify the characteristics of zero inflation in the data sets, as zero inflation has significant impacts on model performance. 
Both datasets meet the characteristics of zero-inflated data whose ratios of zero data are 73.1\% and 88.7\%, respectively. Since the time slot length of traffic accident risk data is much smaller than that of crime data, the traffic accident risk data is more sparse. We make the following observations from the results summarized in Table~\ref{tab:performanceTraffic}.

\begin{table*}[!t]
\renewcommand{\arraystretch}{1.3}
\caption{Performance comparison of \sysname and baselines for traffic accident risk forecasting. RMSE* and MAE* represent the performance on non-zero samples, while RMSE and MAE represent the performance with zero inflation.}
\label{tab:performanceTraffic}
\centering
\begin{tabular}{ccccccccc}
\toprule
\multicolumn{2}{c}{Method}    & \multirow{2}{*}{SVR} & \multirow{2}{*}{MLP} & \multirow{2}{*}{GRU} & \multirow{2}{*}{GMAN} & \multirow{2}{*}{ST-SHN} & \multirow{2}{*}{MAPSED} & \multirow{2}{*}{\sysname} \\
Dataset              & Metric &                      &                      &                      &                       &                         &                         &                       \\
\cmidrule(r){1-9}
\multirow{4}{*}{NYC} & RMSE & 1.1132 & 1.3826 &	1.2565 & 1.2792 & 1.4456 & 1.2702 & \textbf{1.0911} \\
                     & MAE  & 0.8485 & 1.0615 & 0.6921 & 0.7036 & 0.9570 & 0.6754 & \textbf{0.5588} \\
\cmidrule(r){2-9}
                     & RMSE* & 1.7212 & 2.0388 & 1.3360 & 1.3342 & 1.3858 & 1.4209 & \textbf{1.0810} \\
                     & MAE*  & 1.3139 & 1.4887 & 0.7399 & 0.6784 & 0.9225 & 0.8910 & \textbf{0.5663} \\
\cmidrule(r){1-9}
\multirow{4}{*}{CHI} & RMSE & 0.5860 & 0.5852 &	0.6835 & 0.6310 & 0.8544 & 0.6197 & \textbf{0.5682} \\
                     & MAE  & 0.3453 & 0.3434 & 0.4056 & 0.3923 & 0.5412 & 0.3290 & \textbf{0.2581} \\
\cmidrule(r){2-9}
                     & RMSE* & 0.9617 & 0.9741 & 0.6405 & 0.6486 & 0.7447 & 0.7029 & \textbf{0.5749} \\
                     & MAE*  & 0.6801 & 0.7004 & 0.3828 & 0.3604 & 0.4646 & 0.4310 & \textbf{0.2590} \\
\bottomrule
\end{tabular}
\end{table*}

%

\textbf{First}, we observe that \sysname significantly outperforms all baseline models on both datasets and on both metrics. More specifically, \sysname improves the MAE metric by 47.28\% and the RMSE metric by 18.34\% in NYC, respectively. In CHI, \sysname improves the MAE metric by 52.19\% and the RMSE metric by 16.15\%, respectively. This clearly demonstrates the advantages of our model and suggests that our design for \sysname is effective. Especially in the MAE metric, \sysname has an impressive improvement.

\textbf{Second}, we observe differences between different baseline models. 
More specifically, the traditional statistical method SVR and the simple fully connected neural network MLP work better on the more sparse CHI dataset than on the NYC dataset. However, they have large RMSE* and MAE* values. This may be because simple models are more likely to simply predict zero-inflated data as zero, and thus make RMSE and MAE look better in sparser datasets. In fact, they simply predict the number of crimes to be all zeros after flooring~\cite{DBLP:journals/corr/abs-2110-01794}, thus achieving the relatively low RMSE and MAE with high RMSE* and MAE*. That's why SVR and MLP perform even better than some deep learning baselines on the integrated CHI dataset. 
GRU can capture the temporal dependence of time series. GMAN models temporal and spatial dependence respectively. Thus their ability to model spatiotemporal patterns is gradually enhanced on the NYC dataset. However, they are less effective on the CHI dataset, suggesting that they are more suitable for data with a more balanced distribution. 
ST-SHN extracts not only spatiotemporal dependence but also semantic correlation. However, the categories are not distinguished in the traffic accident risk data, so the advantage of ST-SHN is no longer beneficial. At the same time, because the model is specific to the crime prediction task, it performs ineffectively in the traffic accident risk prediction task. 
We also observe that zero inflation does have significant negative impacts on the performance of GRU, GMAN and ST-SHN: their performances of RMSE and MAE on all samples are consistently much worse than those without zero-inflated issues. It reveals that these methods are not good at learning traffic accident risk patterns while demonstrating the importance of solving the zero-inflated problem. 

\textbf{Third}, MAPSED is an exception in that it performs particularly well on zero-inflated samples. However, it only uses deep learning methods in the model that are experimentally more suitable for zero-inflated data, but does not design a dedicated module to solve this issue, which explains its inferior performance compared with \sysname.
Additionally, MLP performs the worst due to the fact that it is not applicable to spatiotemporal data.

\subsection{Model Ablation Study (Q2)}
\begin{figure}[!t]
\centering
\subfloat[MAE]{\includegraphics[width=0.5\linewidth]{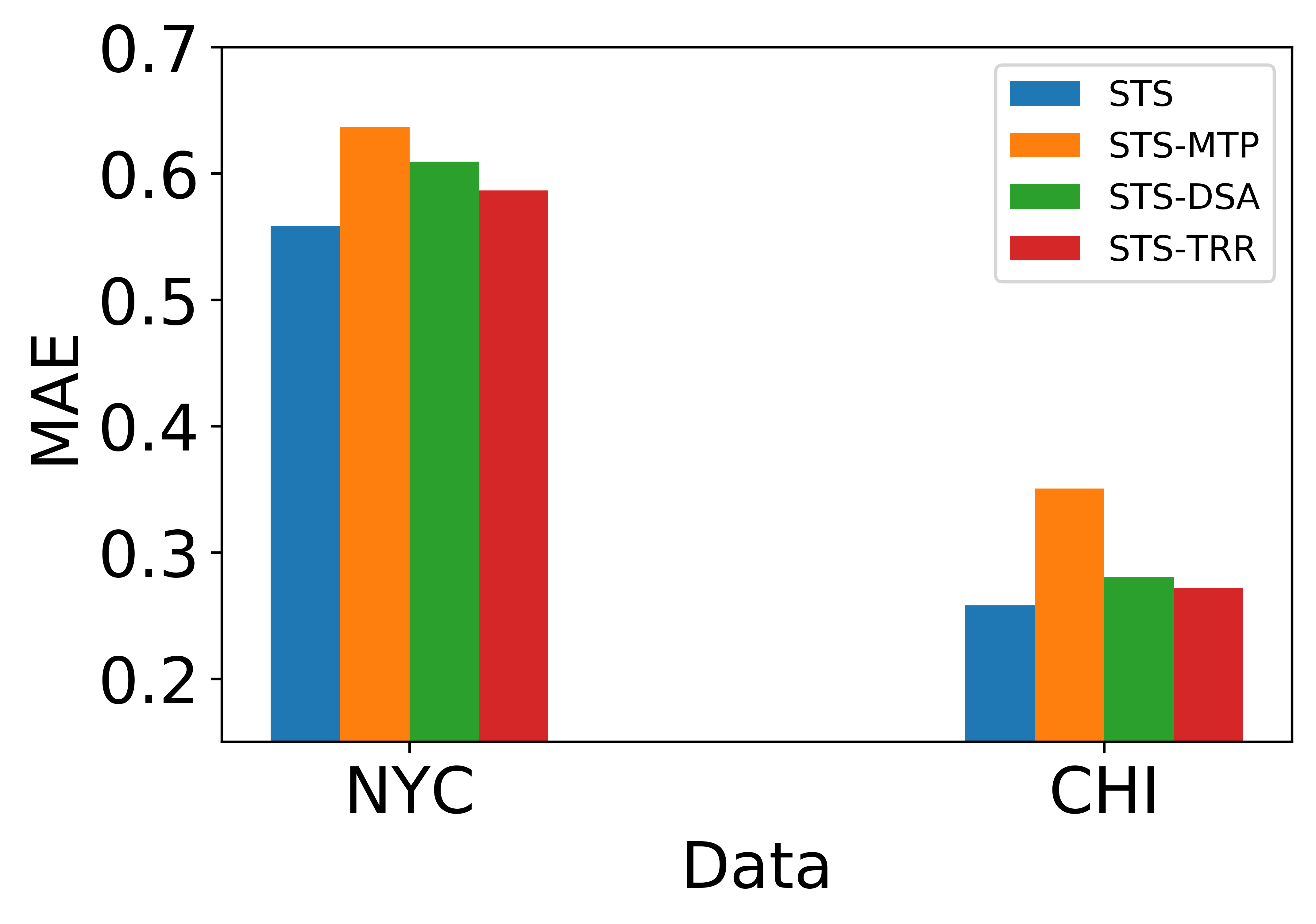}%
\label{fig:mae2}}
\hfil
\subfloat[RMSE]{\includegraphics[width=0.5\linewidth]{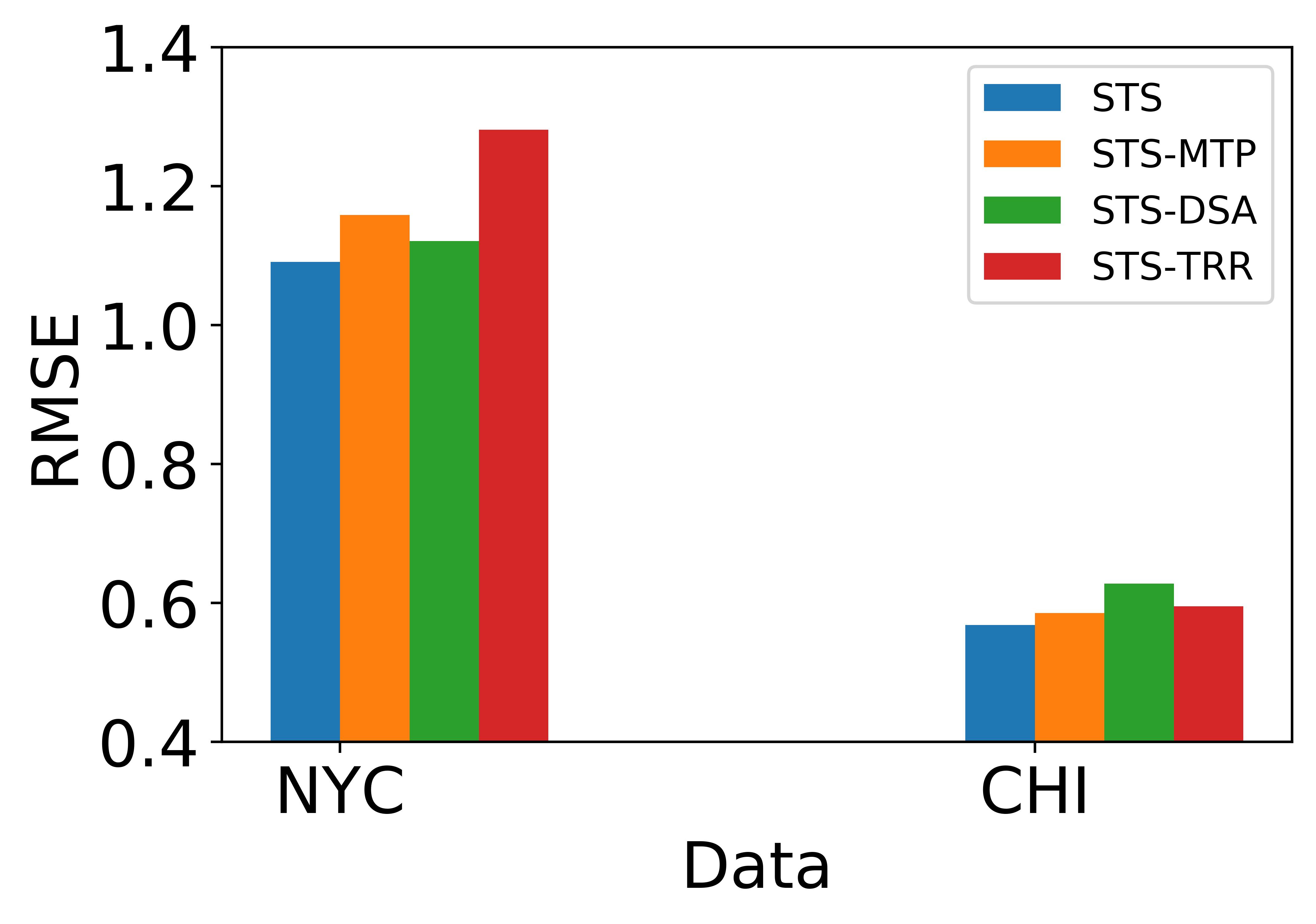}%
\label{fig:rmse2}}
\caption{Model ablation study for traffic accident risk prediction.}
\label{fig:ablationTraffic}
\end{figure}

Because traffic accident datasets do not distinguish between categories, we remove the MSA module in this task. We consider three different variants to evaluate the effectiveness of the components in \sysname: (i)-TRR: we remove the time-aware recurrent module which is used to capture the short-term time dependence; (ii)-DSA: we remove the dynamic spatiotemporal attention in the STC module; (iii)-MTP: instead of the multi-task prediction module which is used to handle the zero-inflated issue, we simply apply a linear regression to predict the result.

We summarize in Figure~\ref{fig:ablationTraffic} the results of model ablation study.
We observe that the integrated \sysname achieves the best performance, which demonstrates all the components are useful.
More specifically, as for the MAE metric, the replacement of the multi-task prediction module has the largest impact, which proves that the proposed multi-task prediction module effectively solves the zero-inflated problem. Removing the DSA and TRR components also impact the performance of \sysname, which demonstrates that simultaneously capturing the temporal, geospatial  and spatial-temporal correlations is critical to traffic accident risk prediction. Among these  variants, the -TRR variant has the worst performance, suggesting that short-term temporal correlation is more relevant with the traffic accident risk prediction task. 
Regarding the RMSE metric, the three variants don't show particularly distinct identical characteristics on both datasets, but these variants are all less effective than the integrated \sysname, indicating the usefulness of each component.

\subsection{Hyperparameters Studies (Q3)}
\begin{figure*}[!t]
\centering
\subfloat[Number of STC Layers]{\includegraphics[width=0.3\textwidth]{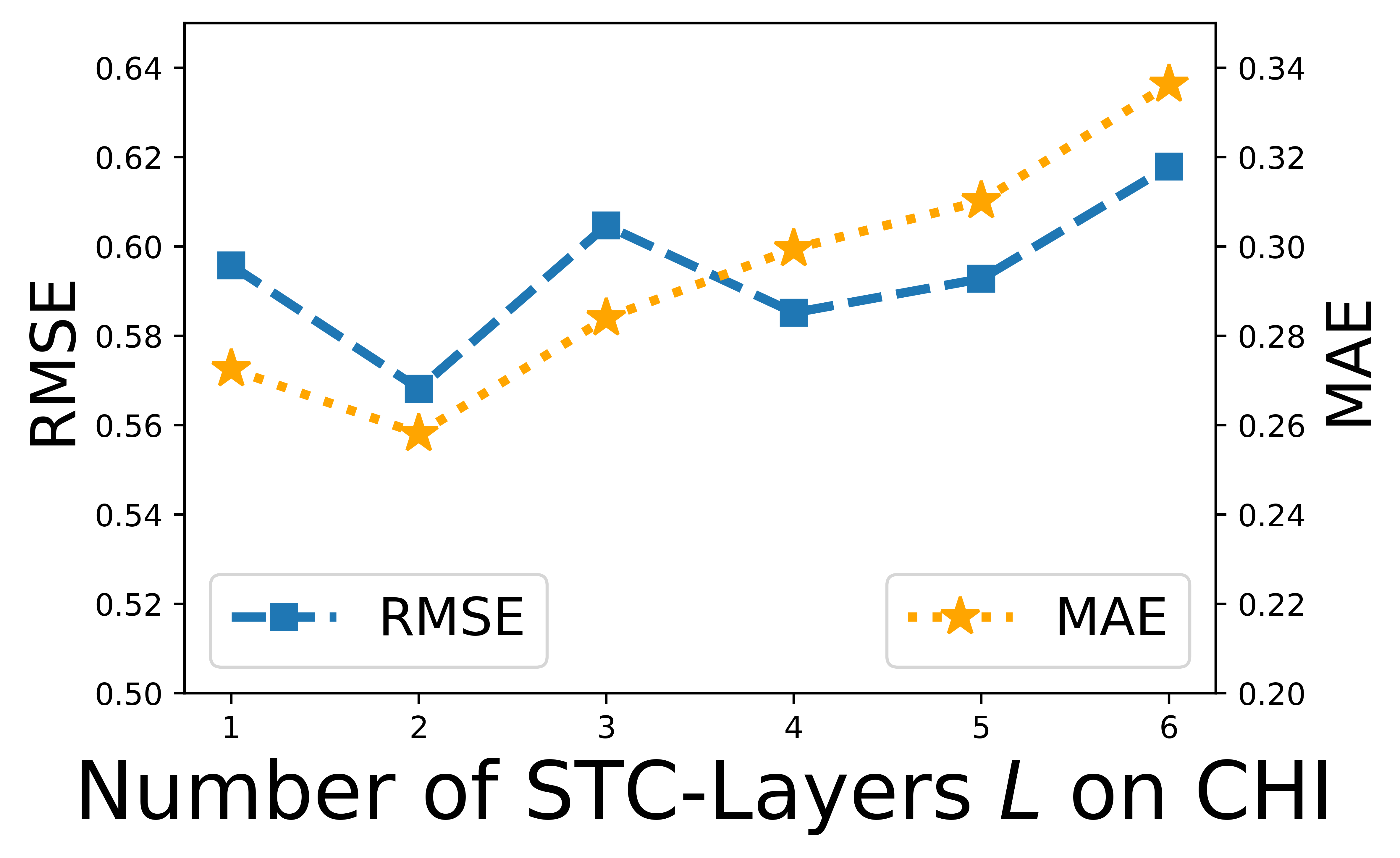}}
\hfil
\subfloat[Number of Head]{\includegraphics[width=0.3\textwidth]{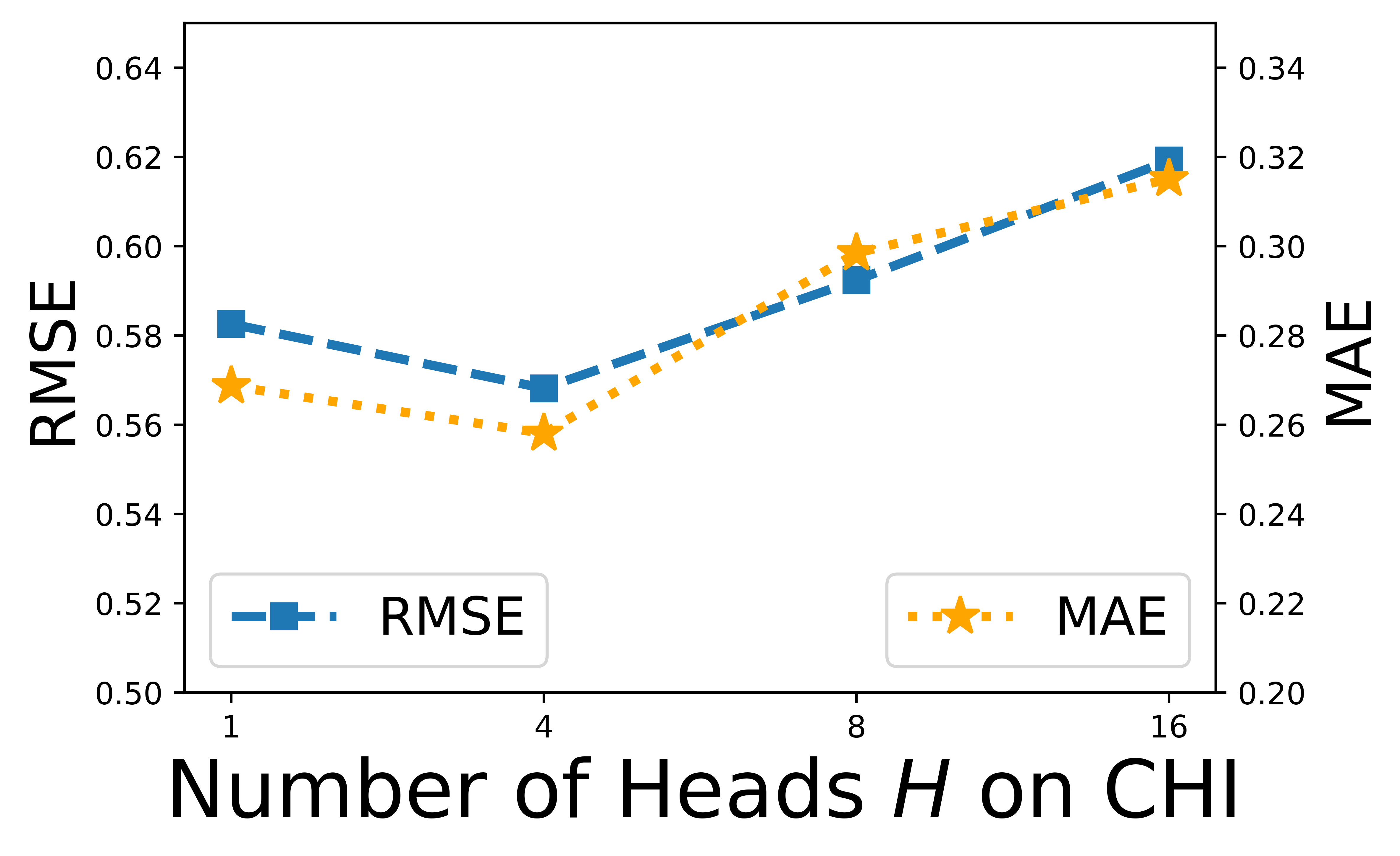}}
\hfil
\subfloat[Hidden State Dimension]{\includegraphics[width=0.3\textwidth]{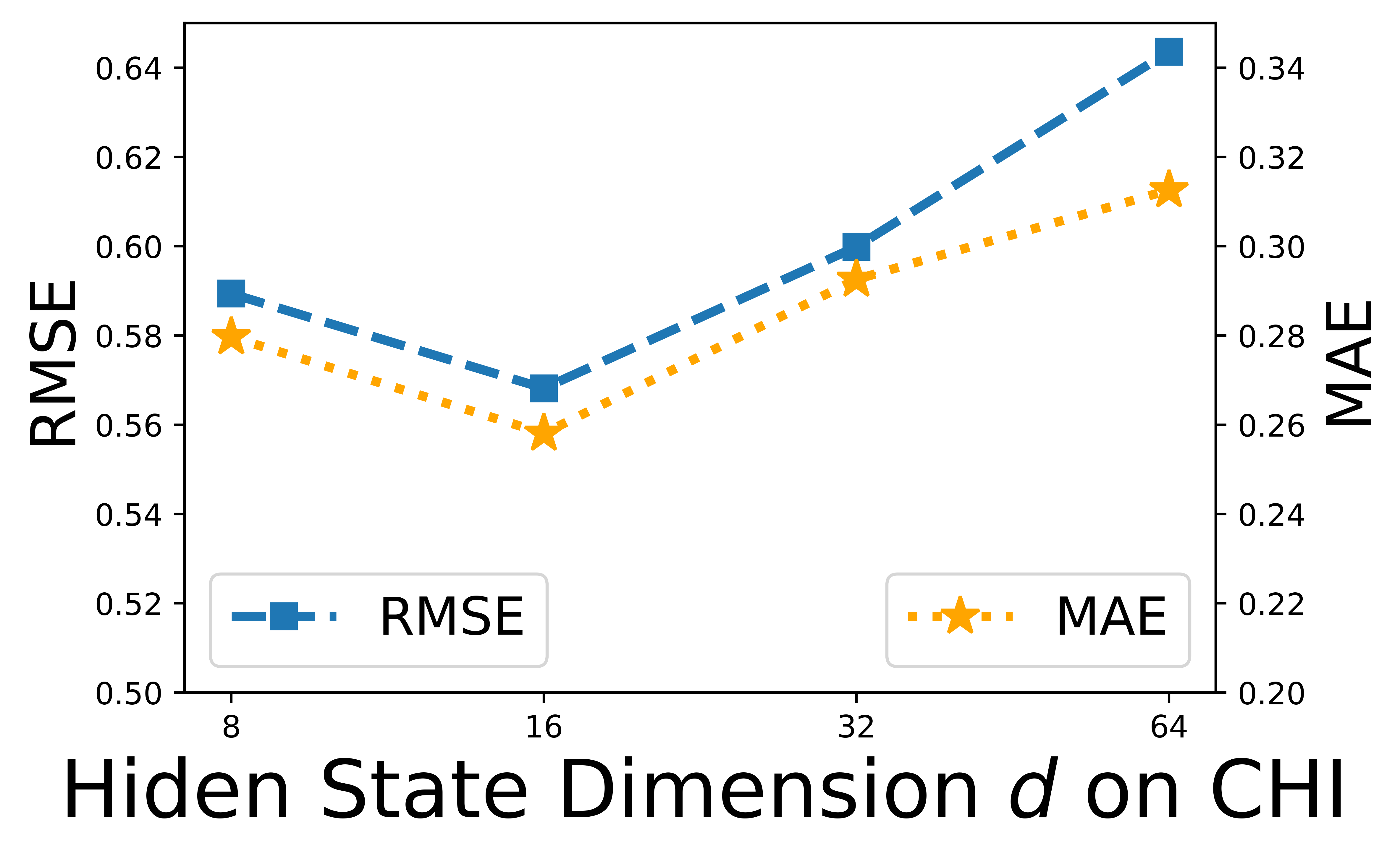}}
\caption{Hyperparameters studies on CHI. Performance on NYC is similar.}
\label{fig:parameterTraffic}
\end{figure*}

%
We summarize the results of the hyperparameters studies on CHI in Figure~\ref{fig:parameterTraffic}.

We make the following observations. 
First, \sysname does not show considerable performance fluctuation during the tuning, which suggests that \sysname is robust.
%
Second, when we gradually vary the number of the STC layers $L$ from 1 to 6, the performance first rises and then falls, with the best performance at $L=2$. This is because with too few layers, the model can only aggregate the influence of neighbors within a few number of hops, thus lacking the strong ability to capture sufficient geospatial correlations. On the other hand, with too many layers, the model absorbs too much noise learned from distant neighbors and thus its performance is degraded. 
Note that a model with too many layers needs to train a much larger number of parameters and is thus prone to become over-fitted. Similarly, when we vary the number of heads in attention mechanism $H$ and the hidden state dimension $d$, it also appears that the performance of \sysname first gets better and then falls back. The reason is, with too few heads or small dimension, the model is under-expressed, while on the contrary the model becomes over-fitted. The best results of \sysname are achieved when $H=4$ and $d=16$, respectively. 
The performance results on NYC are similar and we omit the results due to space limit.


\subsection{Summary}

The evaluation results validate the effectiveness of \sysname on two real-world traffic accident risk zero-inflated datasets. Specifically, \sysname improves on average by 49.74\% and 17.25\% over baselines in MAE and RMSE metrics, respectively. 
We observe that all components of \sysname bring gains. Finally, we experimentally demonstrate the robustness of \sysname and analyze the effect of different settings of hyperparameters on the model.

\section{Takeaway}\label{sec:takeaway}
We conduct extensive experiments on two urban anomaly prediction scenarios (crime prediction and traffic accident risk prediction). The experimental findings described in  Section~\ref{sec:crime} and Section~\ref{sec:traffic} lead us to the following conclusions. 

\textbf{First}, we observe that \sysname performs well on different kinds of urban anomalies, demonstrating its generalizability. Most related methods focus on the characteristics of a specific urban anomaly and perform not so well on other types of urban anomalies. For instance, the crime prediction model ST-SHN outperforms the traffic prediction model GMAN on non-zero samples of crimes, however, 12.40\% worse than GMAN on non-zero samples of traffic accident risks. \sysname performs better on both urban anomalies, which demonstrates its generalizability and ability to learn different characteristics. 

\textbf{Second}, the sparsity rates of the the two use cases, as shown in Table~\ref{tab:StatisticsSparsity}, clearly shows that the traffic accident risk datasets are much sparser than the crime datasets. This raises a challenge in the experiment, as we find that a sparser dataset often contains batches consisting of 0.  
Therefore, we increase the batch size to contain a reasonable and meaningful distribution to allow the model to learn valuable characteristics.
\begin{table}[!t]
  \renewcommand{\arraystretch}{1.3}
  \caption{Statistics on sparsity rate of datasets.}
  \label{tab:StatisticsSparsity}
  \centering
  \begin{tabular}{ccc}
    \toprule
    Case & City & Sparsity Rate\\
    \midrule
    \multirow{2}{*}{Crime} & NYC & 72.7\% \\
                           & CHI & 67.4\% \\
    \midrule
    \multirow{2}{*}{Traffic Accident Risk} & NYC & 73.1\%\\
                                           & CHI & 88.7\%\\
  \bottomrule
\end{tabular}
\end{table}

\textbf{Third}, \sysname provides a greater improvement in traffic accident risk prediction than crime prediction while the traffic accident risk datasets are much sparser. To verify whether \sysname inherently performs better on sparser datasets, we adjust the time interval of the NYC traffic accident risk dataset to obtain datasets with different sparsity rates. To rule out that better results are caused by more zeros, we conduct experiments using the four metrics: RMSE, MAE, RMSE* and MAE*. 
The experimental results as shown in Table~\ref{tab:sparsity} demonstrate that \sysname does have better results on sparser data, thanks to the dedicated multi-task prediction module and customized loss function we design to solve the pervasive zero-inflated issue in urban anomalies data. 

\begin{table}[!t]
  \renewcommand{\arraystretch}{1.3}
  \caption{Performance on NYC traffic accident risk datasets with different time intervals.}
  \label{tab:sparsity}
  \centering
  \begin{tabular}{cccccc}
    \toprule
    Time Interval & Sparsity Rate & RMSE & MAE & RMSE* & MAE* \\
    \midrule
    4 Hours & 79.94\% & 0.9832 & 0.5129 & 1.0023 & 0.5347 \\
    6 Hours & 73.04\% & 1.0911 & 0.5588 & 1.0810 & 0.5663 \\
    8 Hours & 67.51\% & 1.3476 & 0.6873 & 1.2702 & 0.6326 \\
  \bottomrule
\end{tabular}
\vspace{-0.15in}
\end{table}

\section{Conclusion}\label{sec:conclusion}

We propose a novel spatiotemporal and semantic deep learning model \sysname for zero-inflated urban anomaly prediction. Specifically, we design the spatiotemporal and semantic dependency layer to jointly capture the dependencies across spatial, temporal, and semantic dimensions. Besides, we employ a multi-task prediction module with a customized loss function to solve the zero-inflated issue.
Extensive experiments on two representative application scenarios with four real-world datasets demonstrate that zero inflation has significant negative impacts on SOTA models, and that \sysname outperforms SOTA methods in MAE and RMSE by 37.88\% and 18.10\% on zero-inflated datasets, and, 60.32\% and 37.28\% on non-zero datasets, respectively.
%


\section*{Acknowledgment}
This work is supported by the National Key Research and Development Program of China (2021YFC3300502).


\bibliographystyle{IEEEtran}
\bibliography{main}{}










\end{document}